\documentclass[letterpaper, 10 pt, conference]{ieeeconf}  %

\IEEEoverridecommandlockouts                              

\usepackage{bm}
\usepackage{bbm}
\usepackage{amsmath}
\usepackage[amsmath,thmmarks]{ntheorem}
\usepackage{amssymb}
\usepackage{mathtools}
\usepackage[
    bookmarks=false,         
    pdfpagemode=UseNone,     
    colorlinks=true,
    linkcolor=blue,      
    citecolor=blue,       
    urlcolor=blue,        
]{hyperref}
\usepackage{pifont}
\usepackage{cleveref}
\usepackage{multirow}
\usepackage{tikz}
\usepackage{pgfplotstable}
\usepackage{siunitx}
\usepackage{algorithm}
\usepackage{algpseudocodex}
\usepackage{booktabs}
\usepackage{multirow}
\usepackage{xspace}
\usepackage{lipsum}
\usepackage[
    font={footnotesize}
]{caption}
\usepackage{cite}
\usepackage{subfigure}

\newcommand{\bo}{{\bm{o}}}
\newcommand{\bx}{{\bm{x}}}
\newcommand{\ba}{{\bm{a}}}

\newcommand{\Rb}{\mathbb{R}}
\newcommand{\condon}{\,|\,}

\DeclareMathOperator*{\argmin}{arg\,min}

\newcommand{\fw}{CLARE\xspace} 

\title{\LARGE \bf CLARE: Continual Learning for Vision-Language-Action Models \\ via Autonomous Adapter Routing and Expansion
}

\author{
Ralf R\"omer$^{*,1}$ \qquad Yi Zhang$^{*,1}$ \qquad Yuming Li$^{1}$ \qquad Angela P. Schoellig$^{1,2,3}$
\thanks{$^{*}$ Equal contribution.}%
\thanks{$^{1}$ Technical University of Munich, Germany; TUM School of Computation, Information and Technology, Department of Computer Engineering, Learning Systems and Robotics Lab; Munich Institute of Robotics and Machine Intelligence (MIRMI).}
\thanks{$^{2}$ Robotics Institute Germany.}
\thanks{$^{3}$ Munich Center for Machine Learning.}
\thanks{Corresponding email: \tt\small \{ralf.roemer@tum.de\}}
}

\begin{document}

\maketitle

\begin{abstract}

To teach robots complex manipulation tasks, a common approach is to fine-tune a pre-trained vision-language-action model (VLA) on task-specific data.
However, since this recipe updates existing representations, it is unsuitable for long-term operation in the real world, where robots must continually adapt to new tasks and environments while retaining the knowledge they have already acquired.
Existing continual learning methods for robotics commonly require storing previous data (exemplars), struggle with long task sequences, or rely on task identifiers for deployment.
To address these limitations, we propose \fw, a general, parameter-efficient framework for exemplar-free continual learning with VLAs.
\fw introduces lightweight modular adapters into selected VLA modules and autonomously expands the model only where necessary when learning a new task, guided by layer-wise feature similarity.
During deployment, an autoencoder-based routing mechanism dynamically activates the most relevant adapters without requiring task labels.
Through extensive experiments on the LIBERO benchmark and five real-world tasks, we show that \fw achieves high performance on new tasks without catastrophic forgetting of earlier tasks, significantly outperforming even exemplar-based methods.
Code, data, and videos are available at our website: \href{https://tum-lsy.github.io/clare}{tum-lsy.github.io/clare}.
\end{abstract}


\section{Introduction}
Robots deployed in homes, hospitals, or warehouses must operate for long periods while facing ever-changing conditions and task demands.
For instance, a household robot may need to operate a newly purchased appliance, or an assistive robot may meet patients with unfamiliar mobility profiles.
In such settings, robots must continually acquire new skills without sacrificing previously acquired capabilities.
This long-term adaptability, known as continual or lifelong learning~\cite{parisi2019continual}, remains an open challenge in robotics despite decades of research~\cite{thrun1995lifelong, billard2025roadmap, dohare2024loss}.

\begin{figure}[t!]
    \centering
    \includegraphics[width=0.98\linewidth]{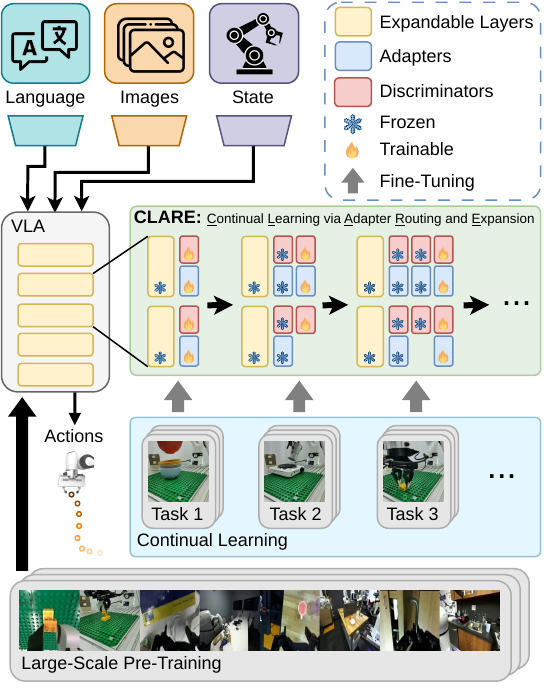}
    \caption{\fw autonomously and continually injects lightweight adapters into selected layers of a pre-trained vision-language-action model~(VLA).
    During inference, the most relevant adapters are activated based on feature similarity, captured by learned discriminators.
    By fine-tuning only the newly added parameters at each stage, the policy acquires new task-specific knowledge without catastrophic forgetting of previously learned skills.
    }
    \label{fig:overview}
\end{figure}

Recent advances in vision-language-action models (VLAs)~\cite{openvla,pi0_5,flower, smolvla} have demonstrated strong performance on complex, long-horizon manipulation tasks by integrating perception, language understanding, and action generation within a unified model.
Pre-training on internet-scale data and robot demonstrations~\cite{khazatsky2024droid, openX} provides VLAs with broad priors that enable some degree of generalization~\cite{pi0_5}.
However, state-of-the-art VLAs still cannot adapt reliably to unseen tasks without fine-tuning on task-specific data~\cite{pi0_5,flower, smolvla}.
In a continual learning setting, where new tasks and environments emerge over time, a naive approach would be to iteratively fully fine-tune~(FFT) a VLA on new data.
However, updating the parameters shared across modalities without regard for previously learned representations leads to significant degradation of both semantic grounding and policy performance on old tasks, a common issue known as catastrophic forgetting~\cite{forgetting}.
Therefore, current VLAs are not inherently capable of continual learning, hindering their deployment in non-stationary, real-world environments.


Experience replay~(ER)~\cite{lopez2017gradient, experience_replay, xie2022lifelong} can mitigate forgetting but has significant practical drawbacks for robotics. Past data may be unavailable due to storage or privacy constraints in a lifelong learning setting. Maintaining and accessing a large replay buffer increases computational and memory overhead, and selecting representative samples for replay is challenging.  
For these reasons, there is a strong need for exemplar-free continual learning methods that enable VLAs to acquire new skills while preserving prior knowledge.

Modular and expandable architectures~\cite{tail,semoa, lei2025dynamic}, which allocate new capacity for each task rather than overwriting shared representations, represent a promising direction for scalable continual learning.
However, these methods often require oracle task identifiers, which are typically unavailable when robots operate autonomously in open-world settings. 
Moreover, the application of these methods to robotics has so far been mostly limited to multitask learning, where all tasks and data are available in advance~\cite{tail,sdp}.

To close this gap, we introduce Continual Learning via Adapter Routing and Expansion~(\fw), a general framework that enables VLAs to continually incorporate new task-specific knowledge without exemplars, task labels, or predefined expansion rules.
\fw injects lightweight adapters into selected modules of the pre-trained model and expands only when the feature statistics indicate substantial novelty, as visualized in~\Cref{fig:overview}.
At deployment, an autoencoder-based routing mechanism dynamically selects among the adapters, enabling autonomous task-agnostic inference. 
This design maintains a balance between stability and plasticity by preserving pre-trained representations and adding capacity as needed, enabling VLAs to learn new tasks with minimal parameter growth.
In summary, our main contributions are:
\begin{itemize}
    \item A lightweight, modular framework enabling VLAs to learn new tasks without catastrophic forgetting.
    \item An autonomous routing mechanism that activates the most suitable adapters during inference using feature similarity without task identifiers.
    \item A dynamic expansion strategy that increases parameter count by only about 2\% per task in our experiments.
    \item Extensive experiments in simulation and real-world settings demonstrating that \fw significantly outperforms continual learning baselines.
\end{itemize}
While our evaluation focuses on a compact VLA architecture, the modular design of CLARE is architecture-agnostic and directly applicable to larger VLAs.
\section{Related Work}
\subsubsection{Vision-Language-Action Models}

In recent years, the robotics community has begun collecting massive multimodal datasets~\cite{openX} and exploiting generative modeling architectures~\cite{ho2020denoising, flow_matching} to equip robots with broad semantic priors and facilitate generalization across tasks and environments~\cite{openvla,flower,pi0_5,smolvla}. 
These VLAs are usually based on vision-language models~\cite{pali3} and pre-trained on robot demonstrations~\cite{openX} via imitation learning, enabling end-to-end mapping from high-dimensional multimodal observations to robot actions.
Despite the growing scale of training data and model capacity, the ability of VLAs to generalize zero-shot to unseen tasks and environments remains limited~\cite{pi0_5, lin2024data}. 
These models often overfit to their pre-training domains, as real-world robot data is much more expensive and scarce than web-scale vision and text datasets~\cite{billard2025roadmap}. 
Thus, it has become a common paradigm to fine-tune a pre-trained VLA on curated, high-quality demonstrations~\cite{openvla, flower, pi0_5, smolvla} to achieve high performance on a specific task.
However, in settings where tasks arrive sequentially and old data may be unavailable, this naive fine-tuning recipe is inadequate, as it overwrites previously learned task knowledge, leading to catastrophic forgetting~\cite{zhou2025learning}.

\subsubsection{Continual Learning} 
Acquiring new skills from a stream of data without catastrophic forgetting of previously learned capabilities or losing plasticity is a hard problem in deep learning~\cite{dohare2024loss}.
ER-based approaches~\cite{lopez2017gradient, experience_replay, xie2022lifelong} retain a subset of past examples and mix them with new data to preserve existing representations during training.
Since storing exemplars may be infeasible, regularization-based methods~\cite{ewc, zenke2017continual} constrain parameter updates for weights deemed important to a past task. 
Related to this idea, PackNet~\cite{packnet} prunes less relevant parameters from previously learned tasks and re-trains them for the new incoming data. 
However, methods like EWC or PackNet struggle with long task sequences as they are restricted by a fixed set of initial parameters.
To overcome capacity bottlenecks and avoid catastrophic forgetting, architectural methods~\cite{rusu2016progressive, tail, semoa} inject new parameters or modules for novel tasks. 
By keeping the original model frozen and leveraging techniques such as low-rank adaptation~(LoRA)~\cite{lora}, these methods can store new task-specific knowledge in a memory-efficient way.


\subsubsection{Continual Imitation Learning in Robotics}
Data scarcity and safety concerns make continual learning for robotics particularly challenging~\cite{thrun1995lifelong, billard2025roadmap}.
LOTUS~\cite{lotus}, a hierarchical method, constructs an ever-growing library of skill policies~\cite{buds} and uses a meta-policy trained using ER for skill selection during deployment.
SDP~\cite{sdp} injects task-specific expert modules into diffusion policies but requires oracle task identifiers during deployment,
which prevents fully autonomous operation in continual learning scenarios.
Another recent work~\cite{ats} fine-tunes a pre-trained base VLA checkpoint for each new task and employs a task scheduler to select from a model library for deployment. 
However, this approach is memory-inefficient and does not enable knowledge sharing between tasks.
MLR~\cite{yu2026lifelong} reduces storage demands by replaying only the frozen-encoder latent embeddings of past demonstrations and regularizes task embeddings via an angular margin constraint to prevent cross-task representation collapse.
Most closely related to our work, DMPEL~\cite{lei2025dynamic} builds an expert library for a transformer-based policy~\cite{libero} and iteratively fine-tunes a router network using replay of the router's past input/output data. 
In contrast to these works, CLARE does not require storing any previous data.

\section{Problem Setup}
We consider a robotic system with state~$\bm{s}_t$ and action~$\ba_t$ at timestep~$t$ that must sequentially learn new tasks~${\{\mathcal{T}_n\}_{n=1}^N}$, with the total number of tasks, $N$, unknown.
A task~${\mathcal{T}_n = (\rho_0^n,\bm{l}_n)}$ is defined by an initial distribution of the state of the robot and the environment~$\rho_0^n$ and a language instruction~$\bm{l}_n$.
We assume the availability of a base VLA policy~${\pi_0 = \pi_{\bm{\theta}_0}}$ with model parameters~$\bm{\theta}_0$ that takes as input an observation~${\bo_t=(\bm{I}_t^1,\dots,\bm{I}_t^{N_\text{c}},\bm{q}_t,\bm{l})}$ consisting of camera images~$\bm{I}_t^{1},\dots,\bm{I}_t^{N_\text{c}}$, proprioceptive state~$\bm{q}_t$ and language command~$\bm{l}$, and generates an action chunk~${\bm{A}_t = (\ba_t,\dots,\ba_{t+H-1}) \sim \pi_0(\cdot\condon\bo_t)}$.
The first~${h\leq H}$ actions in~$\bm{A}_t$ are applied to the robot, and the policy generates a new chunk at timestep~$t+h$ in a receding horizon manner~\cite{chi2023diffusion}.
Pre-training has provided the base VLA with general visual, language, and action representations, but it cannot solve new tasks zero-shot~\cite{pi0_5, flower}.

The robot should be able to learn a new task~$\mathcal{T}_n$ at stage~$n$ while retaining the general knowledge from pre-training and without forgetting how to solve previous tasks~$\mathcal{T}_1,\dots,\mathcal{T}_{n-1}$.
Specifically, given an expert demonstration dataset~${\mathcal{D}_n=\{(\bo_t^n,\bm{a}_t^n), \bm{l}_n\}_{t=1}^T}$ of observation-action pairs for task~$\mathcal{T}_n$, we aim to train a new policy~${\pi_n=\pi_{\bm{\theta}_n}}$ with parameters~${\bm{\theta}_n}$.
For the reasons discussed above, we consider exemplar-free continual learning, i.e., data from earlier stages~$\mathcal{D}_{1},\dots,\mathcal{D}_{n-1}$ are \textit{not} available.
Hence, only the previous model parameters~$\bm{\theta}_{n-1}$ and the new data~$\mathcal{D}_n$ can be used to adapt the policy to the new task.

\section{Methodology}

\begin{algorithm}[tb!]
\caption{Continual learning for VLAs with \fw.}
\begin{algorithmic}[1]
\Require Pre-trained base VLA with parameters~$\bm{\theta}_0$, set of expandable layers~$\mathcal{E}$, expansion threshold~$\gamma$.
\ForAll{layers~$\ell \in \mathcal{E}$}
\State Set~$\mathcal{A}_\ell = \emptyset$, $k_{\ell}=0$. \Comment{Initialize adapter modules} 
\State Set~$\mathcal{D}_\ell = \emptyset$. \Comment{Initialize discriminator modules}
\EndFor
\ForAll{tasks~$\mathcal{T}_n \in \{\mathcal{T}_1,\mathcal{T}_2,\dots\}$}
\Comment{Continual learn.}
\State Set~$\bm{\theta}_n \leftarrow \bm{\theta}_{n-1}$.
\State Collect demonstration data $\mathcal{D}_n$.
\ForAll{layers~$\ell \in \mathcal{E}$}
\Comment{Dynamic Expansion}
\ForAll{discriminators~$D_\ell^j \in \mathcal{D}_\ell$}
\State Compute $z$-score~$z_\ell^j$ via~\eqref{eq:z_score}.
\EndFor
\State Expand~${\mathcal{D}_\ell \leftarrow \mathcal{D}_\ell \cup \big\{D_\ell^n\big\}}$.   
\Comment{New discriminator}
\State Update model parameters: $\bm{\theta}_n \leftarrow (\bm{\theta}_n,D_\ell^n)$.
\If{$n=1$ or $z_\ell^j > \gamma$ for all~$j=1,\dots,{n-1}$}
\State Set~$k_\ell \leftarrow k_\ell + 1$.
\State 
Expand~${\mathcal{A}_\ell \leftarrow \mathcal{A}_\ell \cup \big\{A_\ell^{k_{\ell}}\big\}}$.
\Comment{New adapter}
\State Update model parameters: $\bm{\theta}_n \leftarrow (\bm{\theta}_n,A_\ell^{k_{\ell}})$.
\State Link new discriminator $B_\ell(D_\ell^n) = A_\ell^{k_{\ell}}$.
\Else{}
\State Link~$D_\ell^n$ to an existing adapter via~\eqref{eq:auxiliary_discriminator_linking}.
\EndIf
\EndFor
\If{$n>1$ and no layers~$\ell \in \mathcal{E}$ were expanded}
\State Expand the shallowest layer~$\ell_{1} \in \mathcal{E}$.
\State Re-link~$D_{\ell_1}^n$ to the new adapter.
\EndIf
\State Train all adapters added at stage~$n$ from~$\mathcal{D}_n$ via~\eqref{eq:flow_matching_loss}.
\State Train~$D_\ell^n$ of all layers~$\ell \in \mathcal{E}$  from~$\mathcal{D}_n$ via~\eqref{eq:reconstruction_loss}.
\EndFor
\end{algorithmic}

\label{alg:training}
\end{algorithm}

This section details \fw, our proposed continual learning framework for VLAs.
The training and inference strategies are summarized in Algorithms~\ref{alg:training} and~\ref{alg:routing}.

\subsection{Base Policy} \label{sec:policy}

To learn from high-dimensional, multimodal demonstration datasets, we train the policy using flow matching~\cite{flow_matching}.
The policy at stage~$n$ learns a vector field~$\bm{v}_{\bm{\theta}_n}$ that transports action-chunk samples from a simple base distribution (e.g., a Gaussian) to the target distribution.
We adopt the standard conditional flow matching loss
\begin{align}
    \label{eq:flow_matching_loss}
    \mathcal{L}(\bm{\theta}_n) \!= \mathbb{E}_{s,(\bm{A}^1,\bo),\bm{A}^0}\!\left[\left\|\bm{v}_{\bm{\theta}_n}(\bm{A}^s,\bo,s) - (\bm{A}^1\!-\!\bm{A}^0)\right\|_2\right]\!, \!
\end{align}
where~$s \sim \mathcal{U}([0,1])$, $(\bm{A}^1,\bo)\sim \mathcal{D}_n$, $\bm{A}^0 \sim \mathcal{N}(\bm{0},\bm{I})$, and~$\bm{A}^s = (1-s)\bm{A}^0 + s \bm{A}^1$.
After training, we can generate new action chunks~$\bm{A}_t = \bm{A}^1 \sim \pi_n(\cdot\condon\bo_t)$ by Euler integration of the learned vector field, starting from Gaussian noise~${\bm{A}^0 \sim \mathcal{N}(\bm{0},\bm{I})}$, via~$\bm{A}^{s+\delta s} = \bm{A}^s + \delta s \bm{v}_{\bm{\theta}_n}\left(\bm{A}^s,\bo_t,s\right)$, with~$K=1/{\delta s}$ integration steps.
As our method is architecture-agnostic, we keep the following sections general.

\begin{algorithm}[tb!]
\caption{Autonomous routing during deployment.}
\begin{algorithmic}[1]
\Require Adapters~$\mathcal{A}_\ell$, discriminators~$\mathcal{D}_\ell$, learned linking~$B_\ell:\mathcal{D}_\ell \rightarrow \mathcal{A}_\ell$, input feature~$\bx_\ell$.
\ForAll{discriminators~$D_\ell^j \in \mathcal{D}_\ell$}
\State Compute the reconstruction error~$e_\ell^j(\bx_\ell)$ via~\eqref{eq:reconstruction_error}.
\EndFor
\State Select the most relevant adapter~$A_\ell^* \in \mathcal{A}_\ell$ via~\eqref{eq:routing}.
\State Sum the outputs of original module and adapter via~\eqref{eq:inference}.
\end{algorithmic}
\label{alg:routing}
\end{algorithm}

\subsection{Modularized Adapters} \label{sec:adapters}
The policy must continually acquire new task-specific knowledge, but it should leverage the general representations from pre-training to adapt to new tasks in a parameter-efficient way.
To achieve this, we draw inspiration from the mixture-of-experts~(MoE) approach in large language models~(LLMs)~\cite{shazeer2017outrageously, dai2024deepseekmoe}, which combines the outputs of specialized sub-networks during inference.
However, while their number is fixed in MoE, our setup requires continually injecting parameters into the model to learn new tasks, and we aim to do this in a memory-efficient way.

Prior work~\cite{meng2022locating, geva2021transformer} has shown that a large fraction of factual associations and high-level knowledge in transformer-based LLMs is stored inside mid-layer feedforward network modules.
Motivated by this insight, we define a set of~$n_\text{e}$ expandable modules~${\mathcal{E} = \{\ell_1,\dots,\ell_{n_\text{e}}\}}$ for continual learning.
At each stage~$n$, at most one adapter module is added as a side branch to each expandable layer according to the dynamic expansion strategy detailed in~\Cref{sec:expansion}.
We employ a lightweight encoder-decoder structure with ReLU activation functions for the adapters.
Denoting the input feature of an expandable layer~${\ell \in \mathcal{E}}$ by~$\bm{x}_\ell \in \mathbb{R}^{d_\ell}$, the output of the~$i$-th adapter in that layer is given by
\begin{align}
    A_\ell^i(\bm{x}_\ell) = \bm{W}_{\ell,i}^{\text{up}} \mathrm{ReLU}\big(\bm{W}_{\ell,i}^{\text{down}}\bm{x}_\ell\big), 
\end{align}
where~${\bm{W}_{\ell,i}^{\text{up}} \in \Rb^{d_\ell \times r}}$, ${\bm{W}_{\ell,i}^{\text{down}}\in \Rb^{r \times d_\ell}}$, and~$r \ll d_\ell$.
We denote the set of adapters in layer~$\ell$ at stage~$n$ by~${\mathcal{A}_\ell^n = \big\{A_\ell^1,\dots,A_\ell^{k_{\ell}}\big\}}$ with~$k_{\ell} \leq n$.
To maintain distinct representations for each task, we train only the newly added adapters on the data~$\mathcal{D}_n$ and freeze the rest of the model.
During inference, a routing mechanism described in~\Cref{sec:routing} activates one adapter~${A_\ell^{*} \in \mathcal{A}_\ell}$ per layer, and its output is added to that of the original pre-trained module~$M_\ell^{\text{pre}}(\cdot)$ as
\begin{align}
    \label{eq:inference}
    M_\ell(\bm{x}_\ell) = M_\ell^{\text{pre}}(\bm{x}_\ell) + A_\ell^{*}(\bm{x}_\ell).
\end{align}
Injecting adapters as parallel side branches to the model is beneficial as it preserves the network structure and does not change the input and output of existing layers and adapters.

\subsection{Autonomous Routing} \label{sec:routing}
\begin{figure}[t!]
    \vspace{1.7mm}
    \centering
    \includegraphics[width=\linewidth, trim={0.8cm 0cm 0.3cm 0cm}, clip]{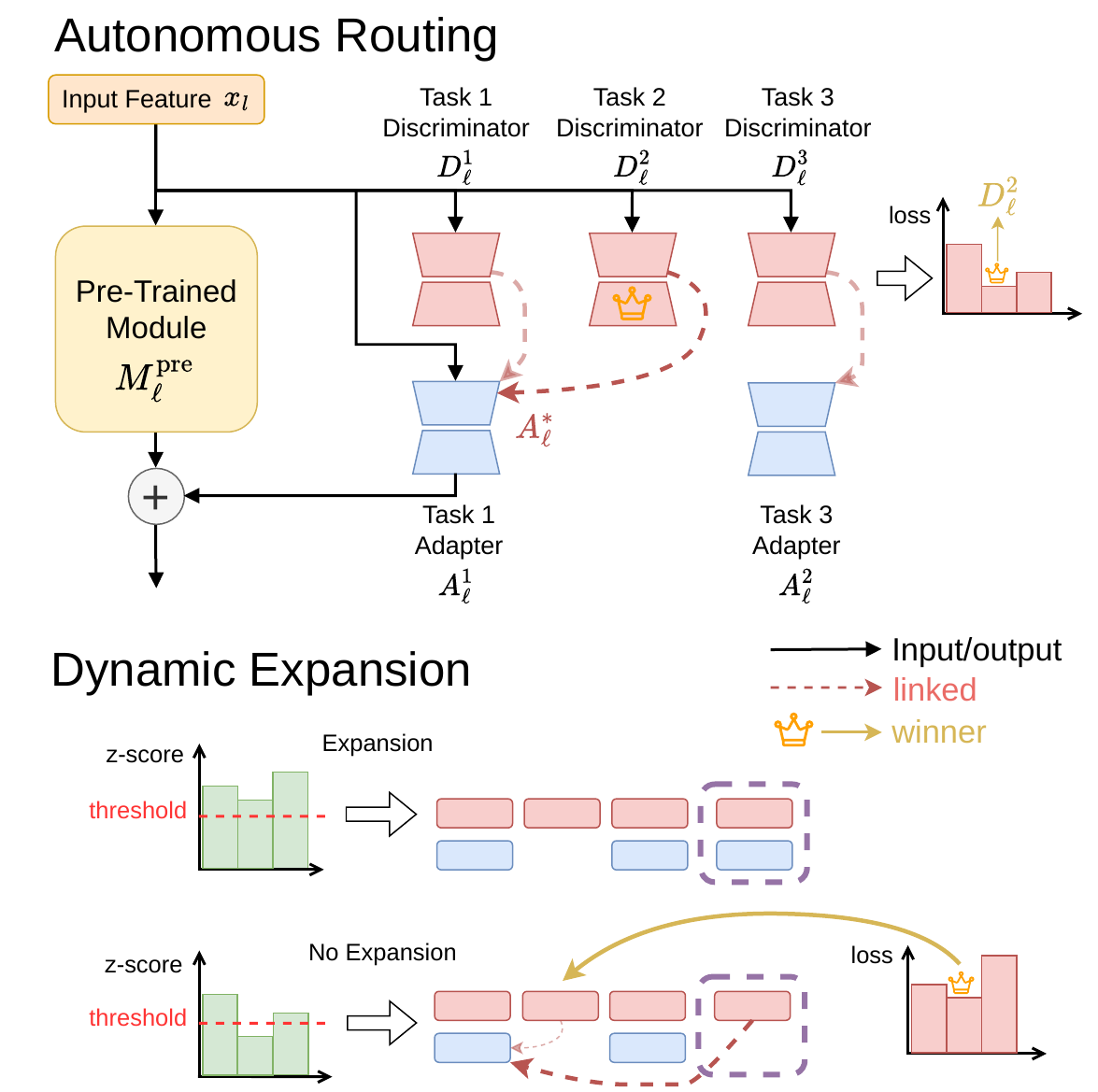}
    \caption{CLARE sequentially adds adapters and discriminators as side branches to selected VLA modules. Top: During inference, our router activates only the most relevant adapter that is linked to the discriminator with the lowest reconstruction error for the input feature. Bottom: During dynamic expansion, if all $z$-scores exceed a threshold~$\gamma$, a new adapter and discriminator are added to the corresponding layer. If at least one $z$-score is smaller than~$\gamma$, we only add a discriminator and link it to the most relevant adapter.}
    \label{fig:dynamic_expansion}
\end{figure}

During deployment, a routing mechanism needs to determine which adapter~$A_\ell^{*} \in \mathcal{A}_\ell$ to activate in each layer~${\ell \in \mathcal{E}}$.
This selection should be autonomous and based solely on the current observation, i.e., without requiring task labels, since these are typically not provided in open, real-world scenarios.
Unlike fixed-size routing in MoE, our setup requires selecting from a continually increasing set of adapters.
We achieve this by designing an expandable and lightweight routing mechanism that selects, for each expandable layer~${\ell \in \mathcal{E}}$, the adapter most relevant to the current situation, as shown in~\Cref{fig:dynamic_expansion}.
We pair each layer with an expanding set of autoencoder discriminators~${D_\ell = \{D_\ell^1,D_\ell^2,\dots\}}$, all of which receive the same features~$\bx_\ell$ as input, and attach a new discriminator~$D_\ell^n$ at each stage~$n$.
Every discriminator~$D_\ell^j$, $j=1,\dots,n$, is linked to one corresponding adapter~${A_\ell^i = B_\ell(D_\ell^j)}$ through a surjective mapping~${B_\ell:D_\ell \rightarrow \mathcal{A}_\ell}$, as explained in~\Cref{sec:expansion}.
We use the reconstruction errors of the discriminators
\begin{align}
    \label{eq:reconstruction_error}
    e_\ell^j(\bx_\ell) = \big\|\bx_\ell - D_\ell^j(\bx_\ell)\big\|_2, \qquad j = 1,\dots,n,
\end{align}
to determine the most relevant adapter.
By training the discriminators added at stage~$n$ with the loss
\begin{align}
    \label{eq:reconstruction_loss}
    \mathcal{L}_\text{recon}(D_\ell^n) = \mathbb{E}_{\bm{x}_\ell \sim \mathcal{D}_n}\big[e_\ell^n(\bx_\ell)\big],
\end{align}
we ensure they have a lower reconstruction error when the input features belong to the training distribution of their corresponding adapter.
During inference, we activate the most relevant adapter linked to the discriminator with the smallest reconstruction error~\eqref{eq:reconstruction_error} via the routing mechanism
\begin{subequations}
\label{eq:routing}
\begin{align}
    A_\ell^*(\bx_\ell) &= B_\ell\big(D_\ell^{j^*}\big), \\
    \text{where} \quad \qquad j^* &= \argmin_{j \in \{1,\dots,n\}} e_\ell^j(\bx_\ell).
\end{align}
\end{subequations}
The routing strategy for one layer is summarized in~\Cref{alg:routing}.
The distribution of features~$\bm{x}_\ell$ changes when training the adapters in the shallower layers.
To nevertheless ensure stable discriminator training, we adopt a two-stage training strategy. First, we jointly train the new adapters using the flow-matching loss~\eqref{eq:flow_matching_loss}. Then, we freeze all parameters except for the new discriminators and train them using the reconstruction loss~\eqref{eq:reconstruction_loss}.
We note that failed demonstrations, if available, could be added to the discriminators' training data to improve the robustness of the routing mechanism.

\subsection{Dynamic Expansion} \label{sec:expansion}
To effectively capture task-specific knowledge without catastrophic forgetting in the context of exemplar-free continual learning, some model expansion is necessary for each new task.
A straightforward approach would be to add new adapters to all expandable layers.
However, this limits knowledge sharing between the tasks and leads to an excessive linear increase in the number of adapter parameters.
Therefore, we only expand a layer~$\ell$ at stage~$n$ if the features of the new task~$\mathcal{T}_n$ deviate substantially from all previous tasks.
Since the discriminators are trained on different data, comparing their reconstruction losses requires normalization.
To this end, we maintain the running mean~$\mu_\ell^j$ and standard deviation~$\sigma_\ell^j$ of the reconstruction loss for each discriminator~$D_\ell^j$ and calculate the normalized~$z$-scores
\begin{align}
    \label{eq:z_score}
    z_\ell^j = \frac{1}{|\mathcal{D}_n|}\sum_{\bx_\ell \in \mathcal{D}_n} \frac{e_\ell^j(\bx_\ell)-\mu_\ell^j}{\sigma_\ell^j}.
\end{align}
If all discriminators~$D_\ell^1,\dots,D_\ell^{n-1}$ have a $z$-score~\eqref{eq:z_score} larger than a threshold~$\gamma$, the features~$\bx_\ell$ of the new task~$\mathcal{T}_n$ in layer~$\ell$ are out-of-distribution with respect to all previously learned tasks.
In this case, we expand the layer by a new adapter~$A_\ell^{k_\ell}$ and link the new discriminator~$D_\ell^n$ to it, i.e., $B_\ell(D_\ell^n)=A_\ell^{k_\ell}$.
Conversely, if at least one discriminator~$D_\ell^j$ yields a $z$-score smaller than~$\gamma$, we deem that the layer~$\ell$ does not require expansion.
This dynamic expansion strategy, illustrated in~\Cref{fig:dynamic_expansion}, results in a memory-efficient, sub-linear increase in the number of adapter parameters. 

Even if a layer is not expanded, we need to attach a new discriminator to it. 
To explain why, we consider the following scenario: 
Assume that at stage~$n$, a new adapter~$A_{\ell_2}^i$ is added to layer~$\ell_2$, but the shallower layer~$\ell_1$ is not expanded. 
Then, the routing mechanism activates only adapters from earlier stages in layer~$\ell_1$ during training of~$A_{\ell_2}^i$.
However, a new adapter~$A_{\ell_1}^j$ could be added to layer~$\ell_1$ at the next stage~$n+1$.
In this case, when revisiting task~$n$, the router might activate the new adapter~$A_{\ell_1}^j$ instead of an earlier one. 
As a consequence, the input features~$\bm{x}_{\ell_2}$ to layer~$\ell_2$ when performing task~$\mathcal{T}_n$ are \textit{different} from those seen during training of~$A_{\ell_2}^i$.
This distribution shift in feature space can lead to unpredictable behavior and task failure~\cite{gu2025safe, romer2025failure}.
To avoid this problem and ensure stable routing, we attach an \textit{auxiliary discriminator} if a layer is not expanded.
The auxiliary discriminator is linked to the same adapter as the existing discriminator with the smallest reconstruction error~\eqref{eq:reconstruction_error}, 
i.e., 
\begin{subequations}
\label{eq:auxiliary_discriminator_linking}
\begin{align}
    \label{eq:auxiliary_discriminator_linking_a}
    B_\ell(D_\ell^n) &= A_\ell^i = B_\ell\big(D_\ell^{j^*}\big), \\
    \text{where} \quad \qquad j^* &= \argmin_{j \in \{1,\dots,n-1\}} \mathbb{E}_{\bx_\ell \sim \mathcal{D}_n} \big[e_\ell^j(\bx_\ell)\big].
\end{align}
\end{subequations}
Intuitively, since adapter~\eqref{eq:auxiliary_discriminator_linking_a} was trained on features most similar to task~$\mathcal{T}_n$, we consider its learned representations to be transferable to~$\mathcal{T}_n$.
We note that with our dynamic expansion strategy, an adapter can potentially be activated by more than one discriminator, as shown in~Figures~\ref{fig:overview} and~\ref{fig:dynamic_expansion}.


{
\setlength{\tabcolsep}{4pt}
\begin{table}[tb!]
    \vspace{1.7mm}
    \centering
    \small
    \begin{tabular}{lcccc}
        \toprule
        & \multicolumn{2}{c}{Adapters} & \multicolumn{2}{c}{Discriminators} \\
        \cmidrule(lr){2-3} \cmidrule(lr){4-5}
        Hyperparameter & Real & Sim. & Real & Sim. \\
        \midrule
        \# Params (linear proj.) & 0.38M & 3.2M  & 6.08M & 1.4M \\
        \# Params (AdaLN)        & 0.75M & 0.26M & 1.00M & 0.33M \\
        Learning rate            & $2\times 10^{-4}$ & $1\times 10^{-4}$ & \multicolumn{2}{c}{$5\times 10^{-4}$} \\
        Learning rate schedule   & \multicolumn{2}{c}{cosine} & \multicolumn{2}{c}{constant} \\
        Batch size               & 256   & 32    & 256   & 32 \\
        Training steps           & \multicolumn{2}{c}{20,000} & 10,000 & 2,000 \\
        Expansion threshold~$\gamma$ & 0.0 & 2.5 & \multicolumn{2}{c}{--} \\
        \bottomrule
    \end{tabular}
    \caption{Model and training hyperparameters. The injected modules are much smaller than the base policy, which has about 200M parameters.}
    \label{tab:hyperparams}
\end{table}
}

We found that introducing at least some new parameters per task is essential for the policy to acquire and retain novel skills. 
In addition, we observed that shallower layers typically exhibit a stronger distribution shift between tasks than deeper layers.
Thus, if no layer is deemed to require expansion, we still add an adapter to the shallowest layer~${\ell_{1} \in \mathcal{E}}$ to capture the peculiarities of a new task. 
At the first stage, we expand all layers~${\ell \in \mathcal{E}}$ by default.
In summary, our dynamic expansion mechanism ensures that \fw adds only a small, task-dependent number of parameters without compromising performance when revisiting previous tasks.

\section{Evaluation}

We conduct extensive simulation and real-world experiments 
with a focus on the following research questions:
\begin{itemize}
    \item \textbf{Q1:} Which layers are best suited for expansion?
    \item \textbf{Q2:} How well can \fw learn new tasks, and is the performance on previous tasks affected?
    \item \textbf{Q3:} Can our autonomous dynamic expansion strategy reuse relevant skills from previous tasks?
    \item \textbf{Q4:} What is the computational overhead of CLARE?
\end{itemize}

\begin{figure}[t!]
    \vspace{1.7mm}
    \centering
    \includegraphics[width=\linewidth]{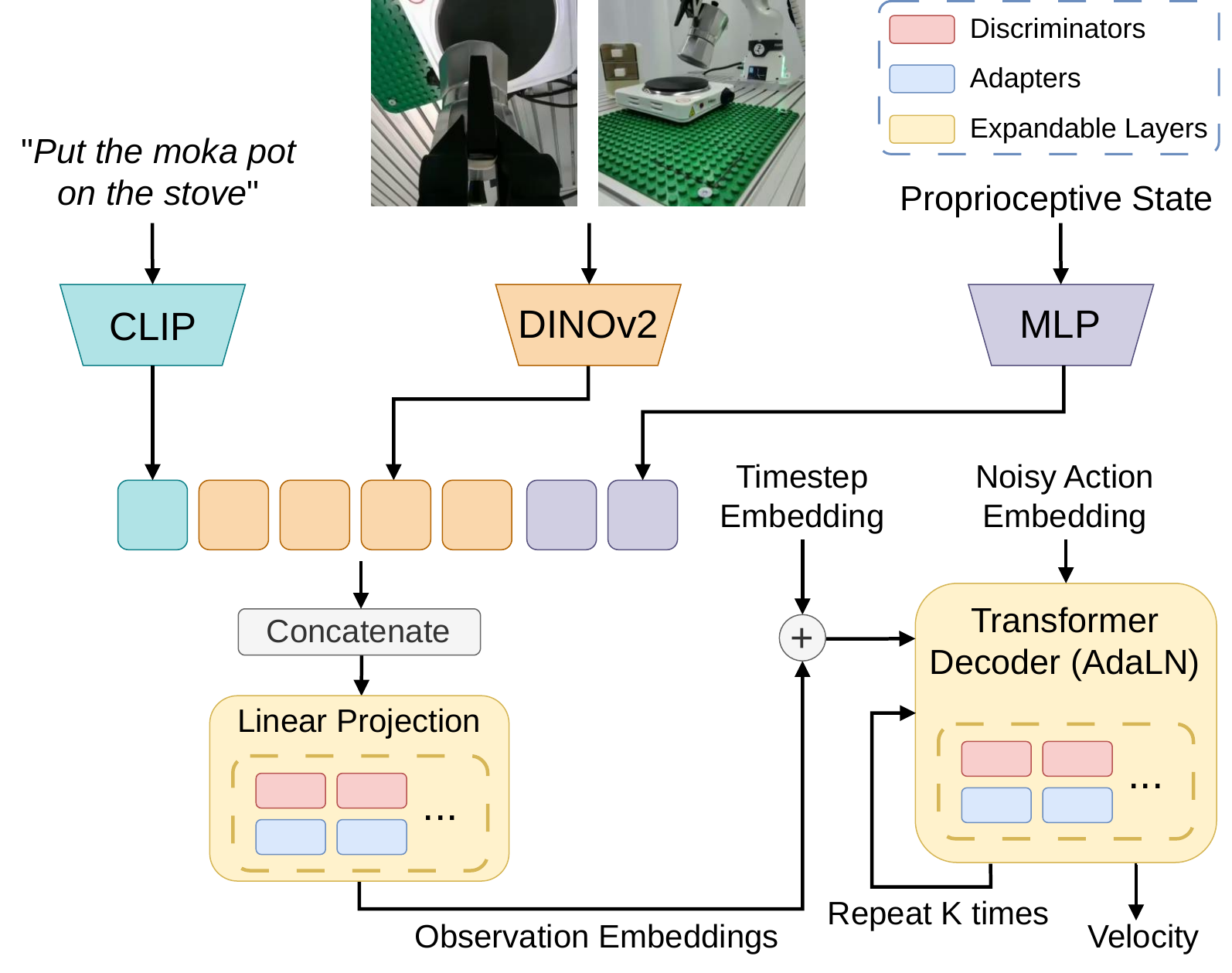}
    \caption{Model architecture of our base VLA policy. 
    The modules for inserting CLARE adapters are shown as dashed blocks. 
    }
    \label{fig:dit_linear}
\end{figure}

\subsection{Experimental Setup}

\subsubsection{Tasks} 
We conduct our simulation experiments using the LIBERO benchmark~\cite{libero}, which is designed specifically for continual learning.
Here, a Panda manipulator with a parallel-yaw gripper needs to perform tasks in a kitchen environment, with 50 human expert demonstrations available per task.
We pre-train the policy on 90 short-horizon tasks from LIBERO-90 and evaluate continual learning on 10 sequentially arriving tasks from LIBERO-Long, LIBERO-Goal, and LIBERO-Spatial. These suites require the robot to understand language instructions and execute various motions, such as pick-and-place, opening a drawer, or turning a knob.

We also conduct hardware experiments with an FR3 manipulator across five tasks, which are visualized in Figure~\ref{fig:tasks}:
\begin{enumerate}
    \item \textsc{Bowl}: Put a bowl on a plate.
    \item \textsc{Stack}: Stack a bowl on top of three other bowls.
    \item \textsc{Moka}: Put a Moka pot on a stove. The pot weighs 0.5$\,\si{kg}$ and hangs at an angle when lifted by its handle.
    \item \textsc{Drawer}: Close a drawer. The friction profile of the plastic drawer necessitates exerting considerable force.
    \item \textsc{Lego}: 
    Put a Lego block into the drawer and close it.
\end{enumerate}
We pre-train the policy on a mix of 1000 demonstrations collected in our lab for tasks different from the five continual learning tasks and 2000 episodes from the DROID dataset~\cite{khazatsky2024droid}.

{
\begin{table}[tb!]
    \vspace{1.7mm}
    \centering
    \begin{tabular}{llccc}
        \toprule
        Backbone & Expandable layers & 
        AUC $\uparrow$ & FWT $\uparrow$ & NBT $\downarrow$ \\
        \midrule
        \multirow{2}{*}{DiT-Dec} & Linear projection & 
        \textbf{75.1}$^{\pm \text{1.3}}$ & \textbf{75.0}$^{\pm \text{1.4}}$ & \textbf{1.9}$^{\pm \text{0.4}}$ \\
        & Decoder & 
        41.8$^{\pm \text{2.4}}$ & 45.5$^{\pm \text{3.8}}$ & 7.0$^{\pm \text{1.7}}$ \\
        \midrule
        \multirow{2}{*}{DiT-EncDec} & Encoder & 
        65.4$^{\pm\text{2.7}}$ & \textbf{66.5}$^{\pm\text{2.2}}$ & 1.7$^{\pm\text{1.2}}$ \\
        & Decoder & 
        29.0$^{\pm\text{2.2}}$ & 30.9$^{\pm\text{4.3}}$ & 3.0$^{\pm\text{3.4}}$ \\
        & Encoder \& Decoder & 
        \textbf{66.6$^{\pm\text{0.3}}$} & 65.8$^{\pm\text{0.4}}$ & \textbf{1.5$^{\pm\text{0.7}}$} \\
        \bottomrule
    \end{tabular}
    \caption{In our ablation study on LIBERO-Long, adding adapters to the observation encoding modules is crucial to achieve strong performance.}
    \label{tab:layer_ablation}
\end{table}
}
\begin{figure}
    \centering
    \includegraphics[width=\linewidth]{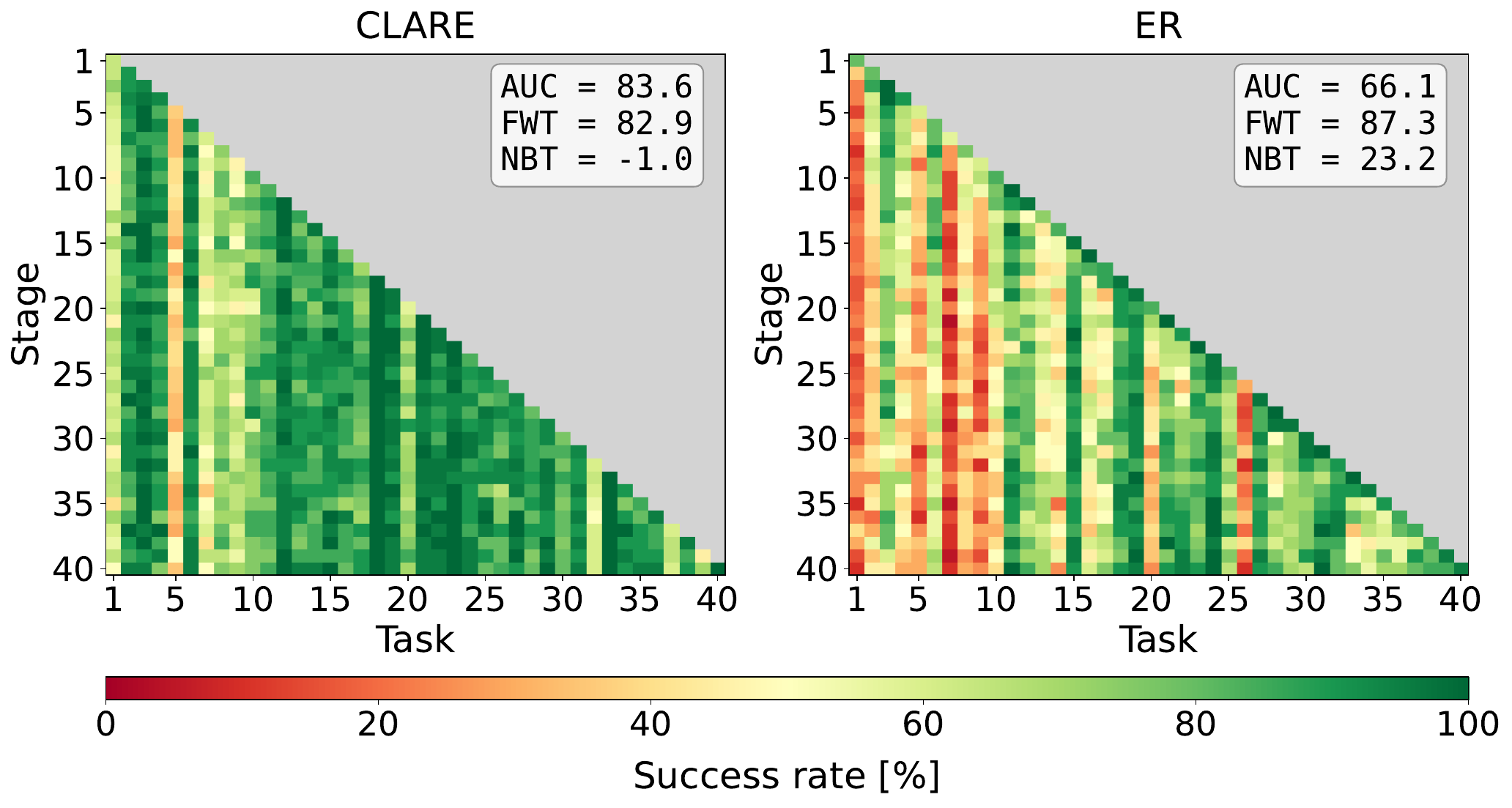}
    \caption{CLARE scales to continual learning of 40 tasks on LIBERO-40, whereas experience replay~(ER) exhibits significant performance degradation.}
    \label{fig:libero_40}
\end{figure}

{
\vspace{1.7mm}
\begin{table*}[bt!]
\centering
\begin{tabular}{l *{9}{r@{}l}}
\toprule
& \multicolumn{6}{c}{LIBERO-Long} & \multicolumn{6}{c}{LIBERO-Goal} & \multicolumn{6}{c}{LIBERO-Spatial} \\
\cmidrule(lr){2-7} \cmidrule(lr){8-13} \cmidrule(lr){14-19}
Method & 
\multicolumn{2}{c}{AUC $\uparrow$} & \multicolumn{2}{c}{FWT $\uparrow$} & \multicolumn{2}{c}{NBT $\downarrow$} & 
\multicolumn{2}{c}{AUC $\uparrow$} & \multicolumn{2}{c}{FWT $\uparrow$} & \multicolumn{2}{c}{NBT $\downarrow$} & 
\multicolumn{2}{c}{AUC $\uparrow$} & \multicolumn{2}{c}{FWT $\uparrow$} & \multicolumn{2}{c}{NBT $\downarrow$} \\
\midrule
SeqFFT         & 22.4 & ${}^{\pm \text{0.3}}$ & \underline{76.1} & ${}^{\pm \text{1.0}}$ & 74.7 & ${}^{\pm \text{1.1}}$ & 26.7 & ${}^{\pm \text{0.9}}$ & \underline{94.1} & ${}^{\pm \text{0.2}}$ & 95.3 & ${}^{\pm \text{1.4}}$ & 27.7 & ${}^{\pm \text{0.6}}$ & \textbf{94.7} & ${}^{\pm \text{0.3}}$ & 94.6 & ${}^{\pm \text{0.9}}$ \\
SeqLoRA        & 21.4 & ${}^{\pm \text{1.0}}$ & 73.1 & ${}^{\pm \text{1.8}}$ & 71.6 & ${}^{\pm \text{1.6}}$ & 26.1 & ${}^{\pm \text{0.3}}$ & 90.1 & ${}^{\pm \text{1.6}}$ & 90.8 & ${}^{\pm \text{1.7}}$ & 27.3 & ${}^{\pm \text{1.3}}$ & 90.1 & ${}^{\pm \text{2.1}}$ & 89.2 & ${}^{\pm \text{1.3}}$ \\
PackNet        & 4.8 & ${}^{\pm \text{0.2}}$ & 37.2 & ${}^{\pm \text{1.0}}$ & 41.3 & ${}^{\pm \text{1.2}}$ & 10.5 & ${}^{\pm \text{0.3}}$ & 60.3 & ${}^{\pm \text{1.0}}$ & 67.0 & ${}^{\pm \text{1.1}}$ & 8.6 & ${}^{\pm \text{0.1}}$ & 54.7 & ${}^{\pm \text{0.5}}$ & 60.3 & ${}^{\pm \text{0.7}}$ \\
ER             & \underline{60.5} & ${}^{\pm \text{0.2}}$ & \textbf{76.6} & ${}^{\pm \text{0.9}}$ & 22.7 & ${}^{\pm \text{1.8}}$ & 76.0 & ${}^{\pm \text{0.9}}$ & \textbf{94.4} & ${}^{\pm \text{0.7}}$ & 25.1 & ${}^{\pm \text{0.5}}$ & \underline{77.6} & ${}^{\pm \text{0.8}}$ & \underline{92.7} & ${}^{\pm \text{1.0}}$ & 20.9 & ${}^{\pm \text{2.0}}$ \\
LOTUS          & 52.9 & ${}^{\pm \text{1.6}}$ & 58.1 & ${}^{\pm \text{0.2}}$ & \textbf{-7.2} & ${}^{\pm \text{3.0}}$ & 56.0 & ${}^{\pm \text{1.0}}$ & 61.0 & ${}^{\pm \text{3.0}}$ & 30.0 & ${}^{\pm \text{1.0}}$ & \multicolumn{2}{c}{NA} & \multicolumn{2}{c}{NA} & \multicolumn{2}{c}{NA} \\
DMPEL          & 58 & ${}^{\pm \text{3}}$ & 55 & ${}^{\pm \text{4}}$ & 7 & ${}^{\pm \text{1}}$ & \underline{78} & ${}^{\pm \text{2}}$ & 68 & ${}^{\pm \text{2}}$ & \textbf{0} & ${}^{\pm \text{1}}$ & 70 & ${}^{\pm \text{3}}$ & 64 & ${}^{\pm \text{2}}$ & \underline{3} & ${}^{\pm \text{1}}$ \\
MLR            & \multicolumn{2}{c}{NA} & \multicolumn{2}{c}{NA} & \multicolumn{2}{c}{NA} & 77.2 & ${}^{\pm \text{1.8}}$ & 80.0 & ${}^{\pm \text{2.5}}$ & 6.9 & ${}^{\pm \text{0.9}}$ & \multicolumn{2}{c}{NA} & \multicolumn{2}{c}{NA} & \multicolumn{2}{c}{NA} \\
\textbf{\fw (ours)} & \textbf{75.1} & ${}^{\pm \text{1.3}}$ & 75.0 & ${}^{\pm \text{1.4}}$ & \underline{1.9} & ${}^{\pm \text{0.4}}$ & \textbf{89.3} & ${}^{\pm \text{1.1}}$ & 89.7 & ${}^{\pm \text{1.5}}$ & \textbf{0.3} & ${}^{\pm \text{1.1}}$ & \textbf{87.4} & ${}^{\pm \text{2.3}}$ & 88.0 & ${}^{\pm \text{1.9}}$ & \textbf{0.9} & ${}^{\pm \text{0.6}}$ \\
\bottomrule
\end{tabular}
\caption{Baseline comparison across three LIBERO suites. \fw achieves the highest overall performance, as measured by AUC, and demonstrates strong capabilities to acquire new skills without forgetting. ``NA'' indicates not available.}
\label{tab:baseline_comparison}
\end{table*}
}

\subsubsection{Policy} 
The observations contain RGB images from wrist and third-person cameras, the end-effector pose and gripper state, and a language command.
The policy generates chunks of~$H=16$ end-effector displacement actions, and the first~$h=8$ actions are sent to a Cartesian controller~\cite{pro2026crisp} at a frequency of~15$\,\si{\Hz}$ (20$\,\si{\Hz}$ in simulation).
For our pre-trained base VLA, we adopt a decoder-only diffusion transformer~(DiT-Dec)~\cite{dit_policy} architecture with 6 transformer layers and adaptive layer normalization~(AdaLN)~\cite{dit} conditioning, as visualized in~\Cref{fig:dit_linear}.
We leverage pre-trained DINOv2~\cite{dinov2} and CLIP~\cite{clip} vision and text encoders, freezing them during continual learning. 
The visual and language features, as well as the proprioceptive state, are first projected into tokens of the same dimension via linear layers before being fed into the transformer backbone.
Adapters can be added to the linear projection layer of the encoder and the transformer layers in the decoder.
As an ablation, we also consider an encoder-decoder backbone~(DiT-EncDec), for which adapters can be added to all 12 transformer layers. 
Our training hyperparameters are provided in~\Cref{tab:hyperparams}.
Training takes about one hour per simulation task and five hours per real-world task on an NVIDIA RTX 5090 GPU.

\subsubsection{Metrics} 
We use three metrics to assess continual learning~\cite{diaz2018don, libero}: Area under the success rate curve (AUC), forward transfer~(FWT), and negative backward transfer (NBT).
Denoting the success rate on task $n$ after learning the first $m\geq n$ tasks as~$r_{n|m}$, the metrics are defined as~$\mathrm{AUC} = \frac{1}{N} \sum_{n=1}^{N} \left( \frac{1}{N - n + 1} \sum_{m=n}^{N} r_{n|m} \right)$, $\mathrm{FWT} = \frac{1}{N} \sum_{n = 1}^{N}{r_{n|n}}$, and~$\mathrm{NBT} = \frac{1}{N - 1} \sum_{n = 1}^{N-1} \left( \frac{1}{N - n} \sum_{m=n+1}^{N} \left(r_{n|n} - r_{n|m}\right) \right)$.
%
%
Intuitively, AUC measures overall performance on new and old tasks, FWT quantifies the ability to learn new tasks, and NBT measures the degree of forgetting (lower being better).
All numerical results are given in percentage points.

\subsubsection{Evaluation Protocol}
After each stage~$n$, we evaluate the policy on all previously learned tasks~$\mathcal{T}_1,\dots,\mathcal{T}_n$ using~100 rollouts per task (10 in the real world), with different initial configurations of relevant objects.
We average all simulation results across three random seeds.

\subsubsection{Baselines} 

We include seven baselines for continual learning without oracle task IDs. \textbf{SeqFFT}~\cite{libero, llm_rlhf} sequentially fine-tunes the whole model for each new task. \textbf{SeqLoRA}~\cite{lora} adds LoRA adapters to selected layers, which are merged back into the base model weights after training on a new task. We add LoRA adapters to all linear layers within the projection, attention, feedforward, and AdaLN modules.
\textbf{PackNet}~\cite{packnet} freezes the most important 25\% of model weights, and the remaining free weights are used to learn the next task. This process is repeated iteratively, with each new task having fewer free weights available.
\textbf{ER}~\cite{experience_replay} trains the model on a mix of 50\% previous and 50\% new data for each new task.
\textbf{LOTUS}~\cite{lotus} maintains a growing skill library and uses a meta policy to compose skills during deployment.
\textbf{DMPEL}~\cite{lei2025dynamic} builds an expert library and iteratively fine-tunes the router using expert coefficient replay.
\textbf{MLR}~\cite{yu2026lifelong} replays compact joint latent representations and regularizes task embeddings to preserve inter-task distinctiveness.
  

\subsection{Simulation Results}

To examine the impact of the set of layers to expand~$\mathcal{E}$ for LIBERO-Long, we set~${\gamma=0}$, i.e., an adapter is added to each expandable layer per new task.
As shown in~\Cref{tab:layer_ablation}, expanding only the encoder part of the model outperforms expanding only the decoder.
This indicates that the
policy's observation-conditioning modules, which perform task-specific modulations of the decoder's action priors, are well-suited to store new knowledge during continual learning.
Hence, we adopt this expansion strategy in the subsequent experiments.


Our baseline comparison is summarized in~\Cref{tab:baseline_comparison}.
For DMPEL and MLR, we report the authors' results.
CLARE achieves the highest overall performance, as measured by AUC, outperforming the best baseline, ER, by about 10 to 14 percentage points.
Compared to SeqFFT and ER, which fine-tune the full model, \fw achieves comparable FWT, indicating that it can store new task-specific knowledge in much fewer parameters.
Moreover, our method achieves approximately zero NBT, demonstrating that it can avoid forgetting without relying on exemplar data or oracle task identifiers. 


We examine long-term scalability on LIBERO-40, which concatenates all four standard LIBERO suites (LIBERO-Long, -Goal, -Spatial, -Object).
Due to computational constraints, we perform only 10 rollouts per task and stage across three seeds.
As shown in~\Cref{fig:libero_40}, CLARE can sequentially learn and retain 40 distinct tasks, demonstrating the scalability and robustness of our autonomous routing strategy.
In contrast, ER cannot avoid catastrophic forgetting of several tasks (e.g., $\mathcal{T}_1$ and $\mathcal{T}_7$), yielding an NBT of 23\%.

\begin{figure}[tb!]
    \vspace{1.7mm}
    \centering
    \includegraphics[width=\linewidth]{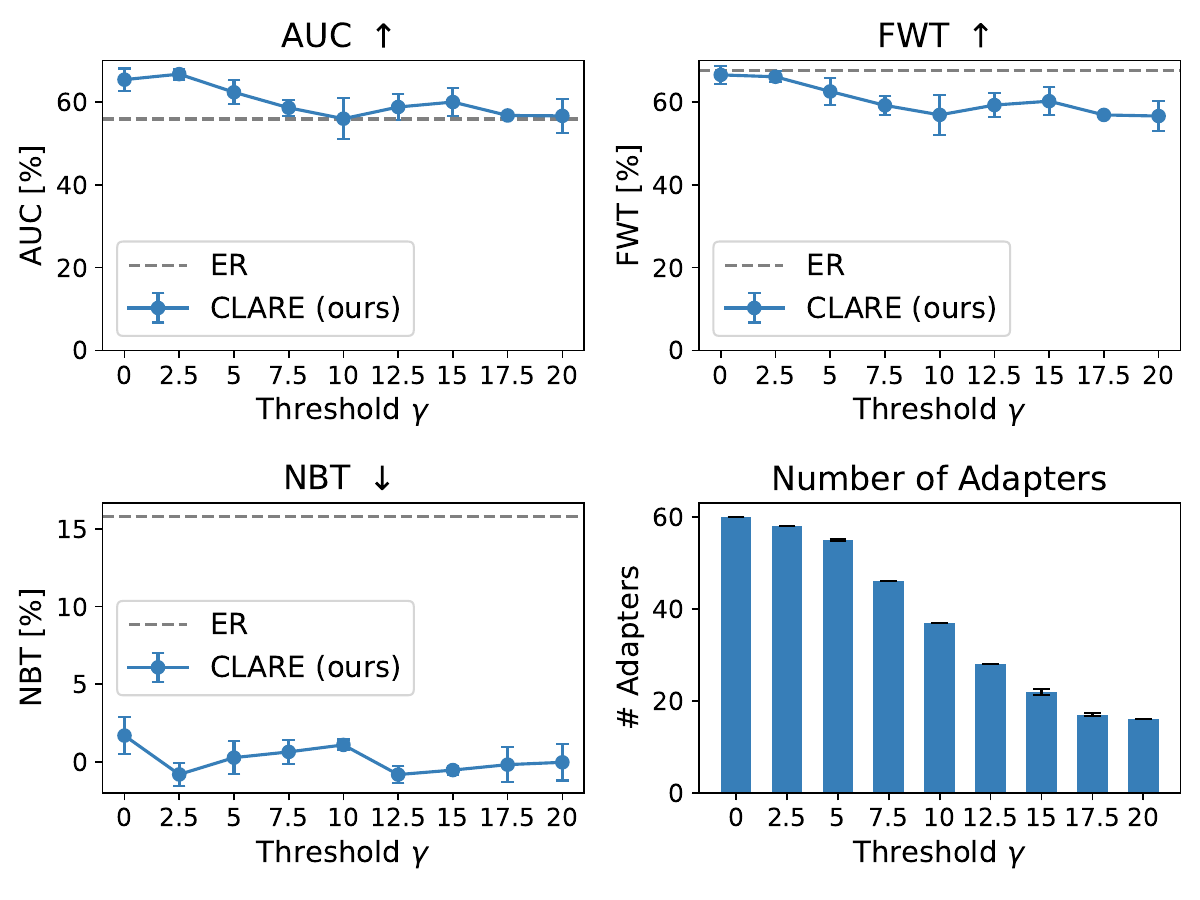}
    \caption{
    Increasing the dynamic expansion threshold~$\gamma$ reduces the number of added adapters and, consequently, the capability to learn new tasks (lower FWT), but does not lead to catastrophic forgetting (approximately zero NBT).}
    \label{fig:threshold_ablation}
\end{figure}


We evaluate the impact of the expansion threshold~$\gamma$ for DiT-EncDec with expandable encoder layers and report the results in~\Cref{fig:threshold_ablation}. 
Increasing~$\gamma$ from 0 to 20 
reduces the number of added adapters from 60 to 16, resulting in a slight decrease in AUC and FWT.
Intuitively, compressing new knowledge into fewer model parameters reduces the robot's ability to learn novel tasks.
Yet, AUC remains higher than for ER, and NBT stays close to zero, indicating that the policy does not exhibit forgetting even when only a few adapters are added per task.
Comparing the results in~\Cref{tab:layer_ablation} and~\Cref{fig:threshold_ablation}, we find that the choice of expandable layers can have a stronger impact than the expansion threshold value.
In summary, $\gamma$ should be chosen based on the importance of high task performance versus low memory requirements.

\renewcommand{\subfigcapskip}{-5pt}
\begin{figure}[tb!!]
\vspace{1.7mm}
    \centering
    \subfigure{%
        \centering
        \includegraphics[width=0.195\linewidth, trim={6cm 0cm 10cm 0cm}, clip]{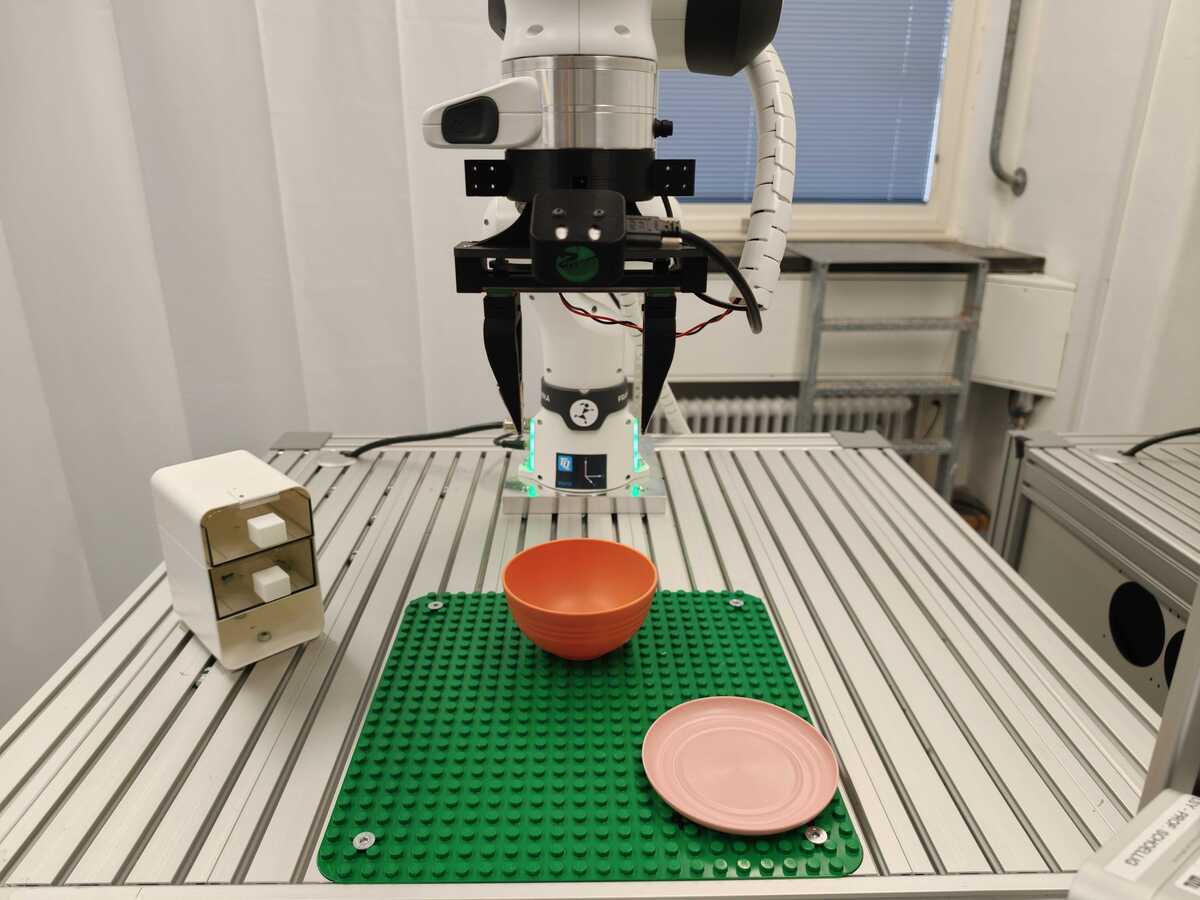}%
    }\hfill
    \subfigure{%
        \centering
        \includegraphics[width=0.195\linewidth, trim={6cm 0cm 10cm 0cm}, clip]{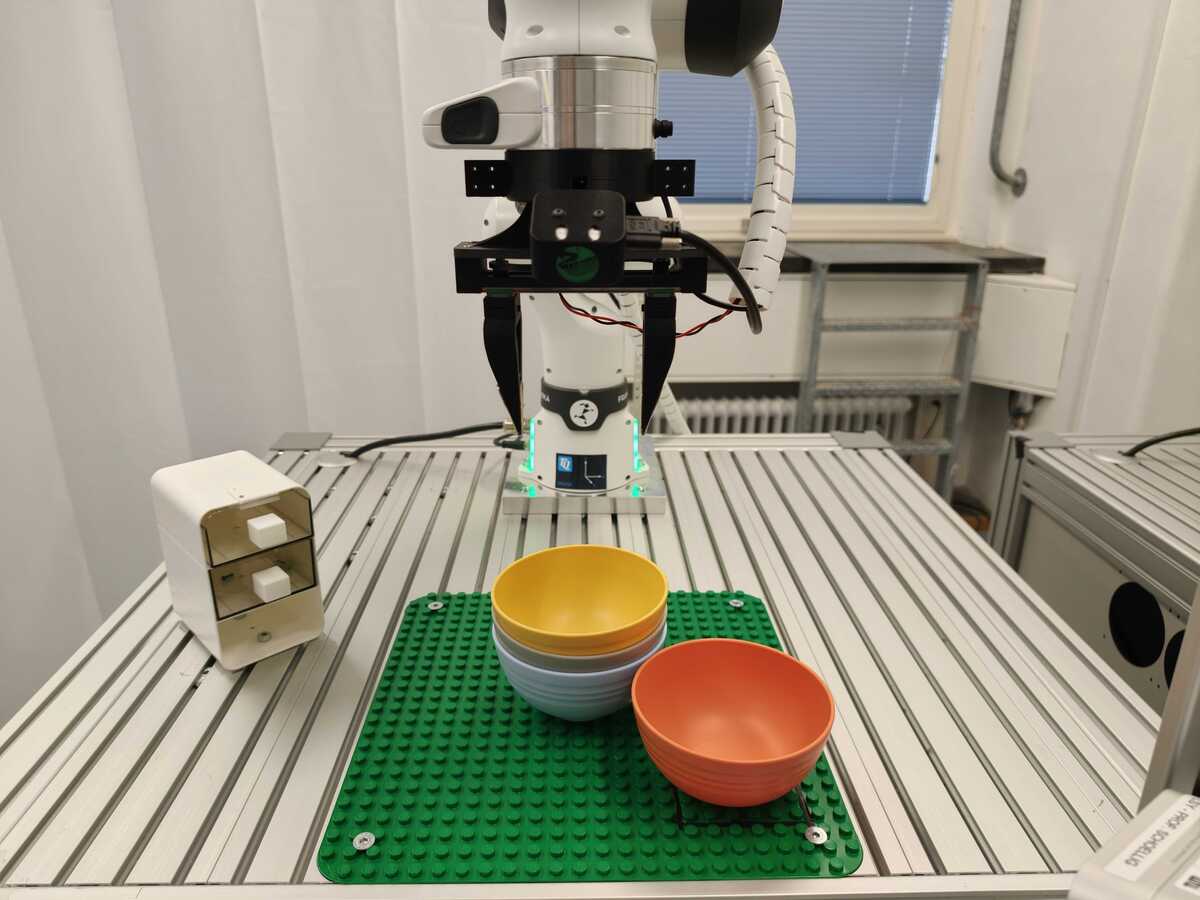}%
    }\hfill
    \subfigure{%
        \centering
        \includegraphics[width=0.195\linewidth, trim={6cm 0cm 10cm 0cm}, clip]{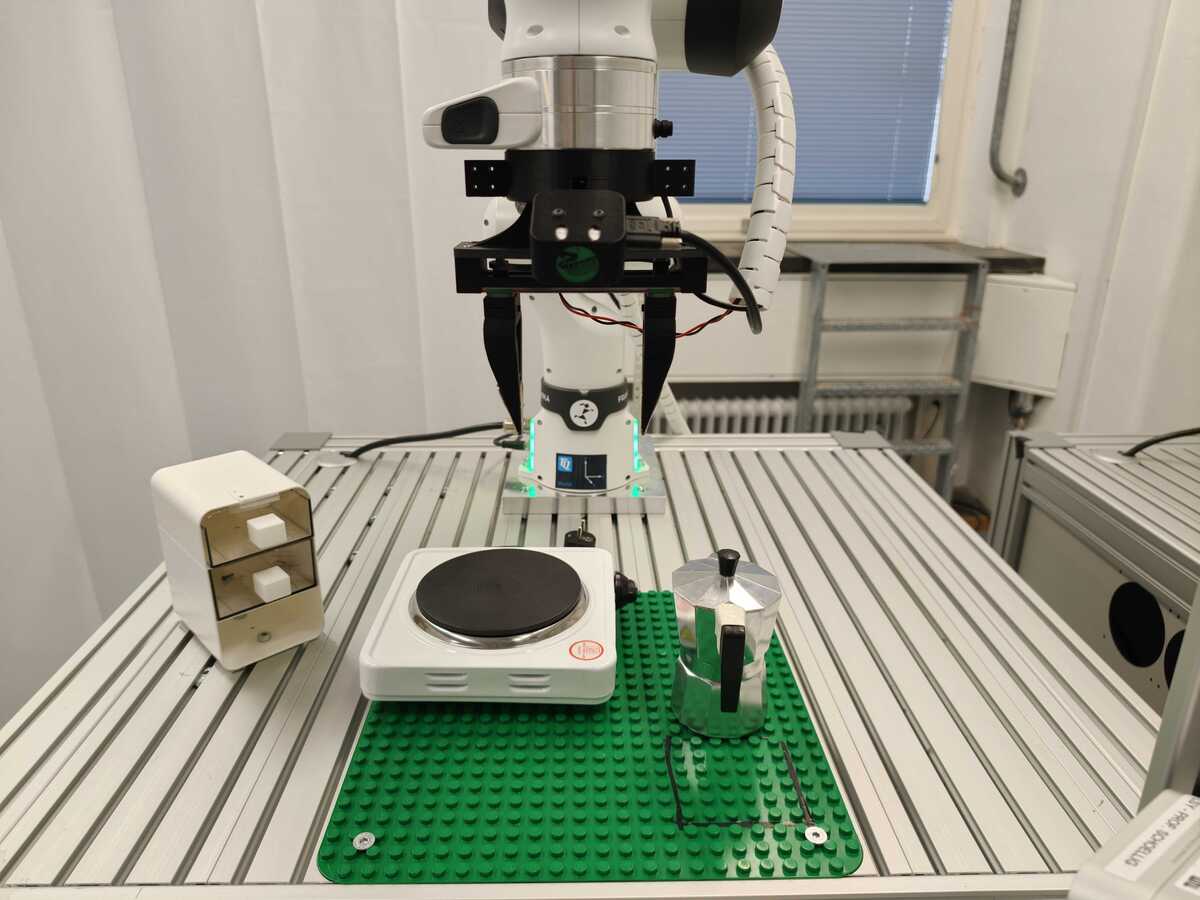}%
    }\hfill
    \subfigure{%
        \centering
        \includegraphics[width=0.195\linewidth, trim={6cm 0cm 10cm 0cm}, clip]{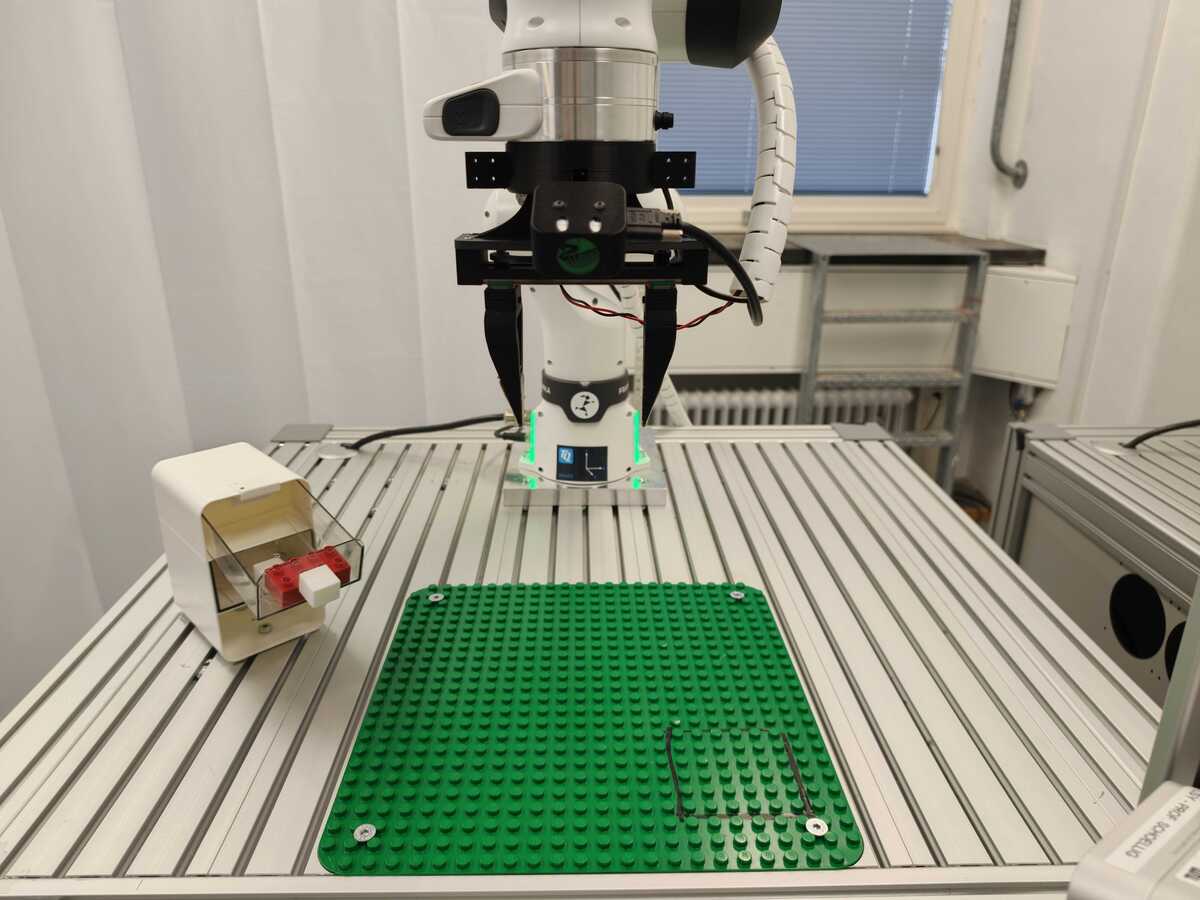}%
    }\hfill
    \subfigure{%
        \centering
        \includegraphics[width=0.195\linewidth, trim={6cm 0cm 10cm 0cm}, clip]{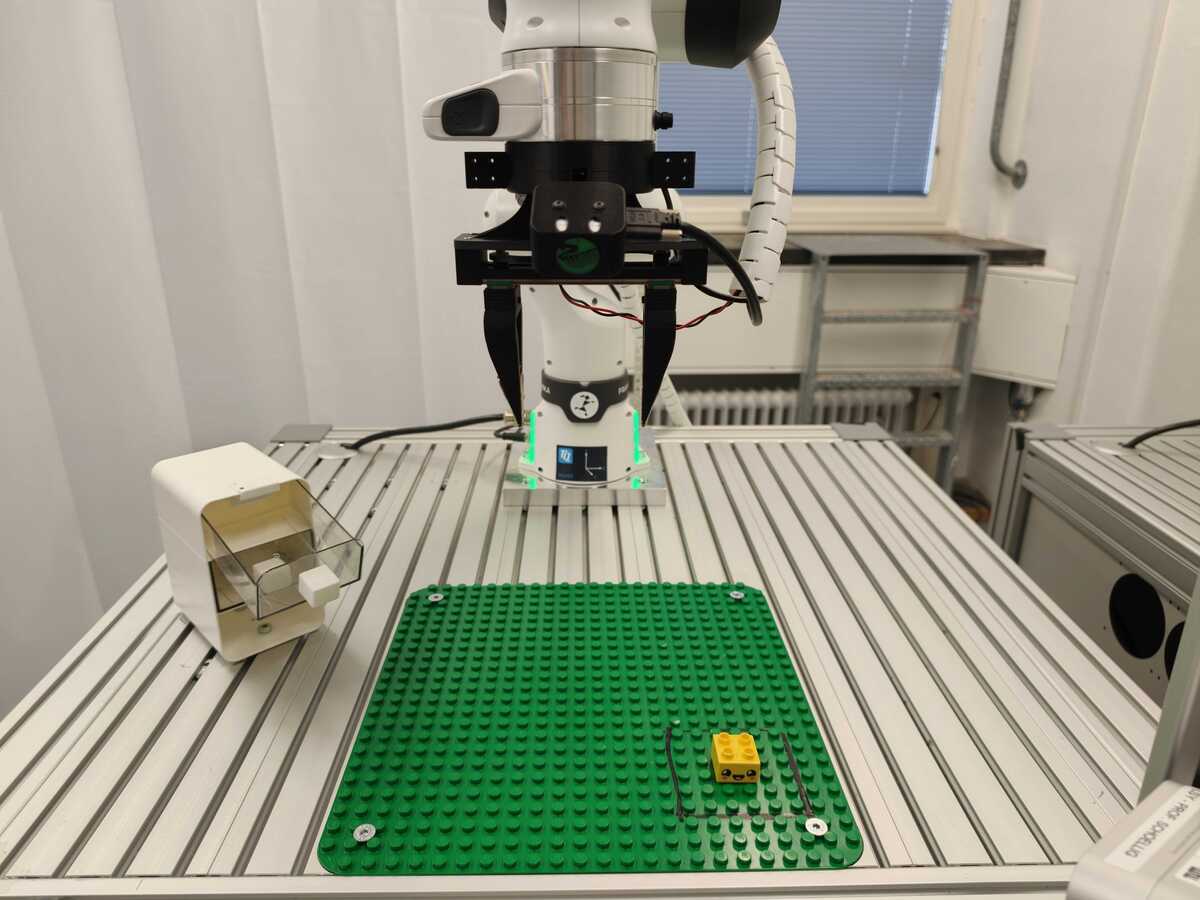}%
    }\hfill\\
    \vspace{-9pt}
    \renewcommand{\thesubfigure}{}
    \setcounter{subfigure}{0}
    \subfigure[1. \textsc{Bowl}
    ]{%
        \centering
        \includegraphics[width=0.195\linewidth, trim={6cm 0cm 10cm 0cm}, clip]{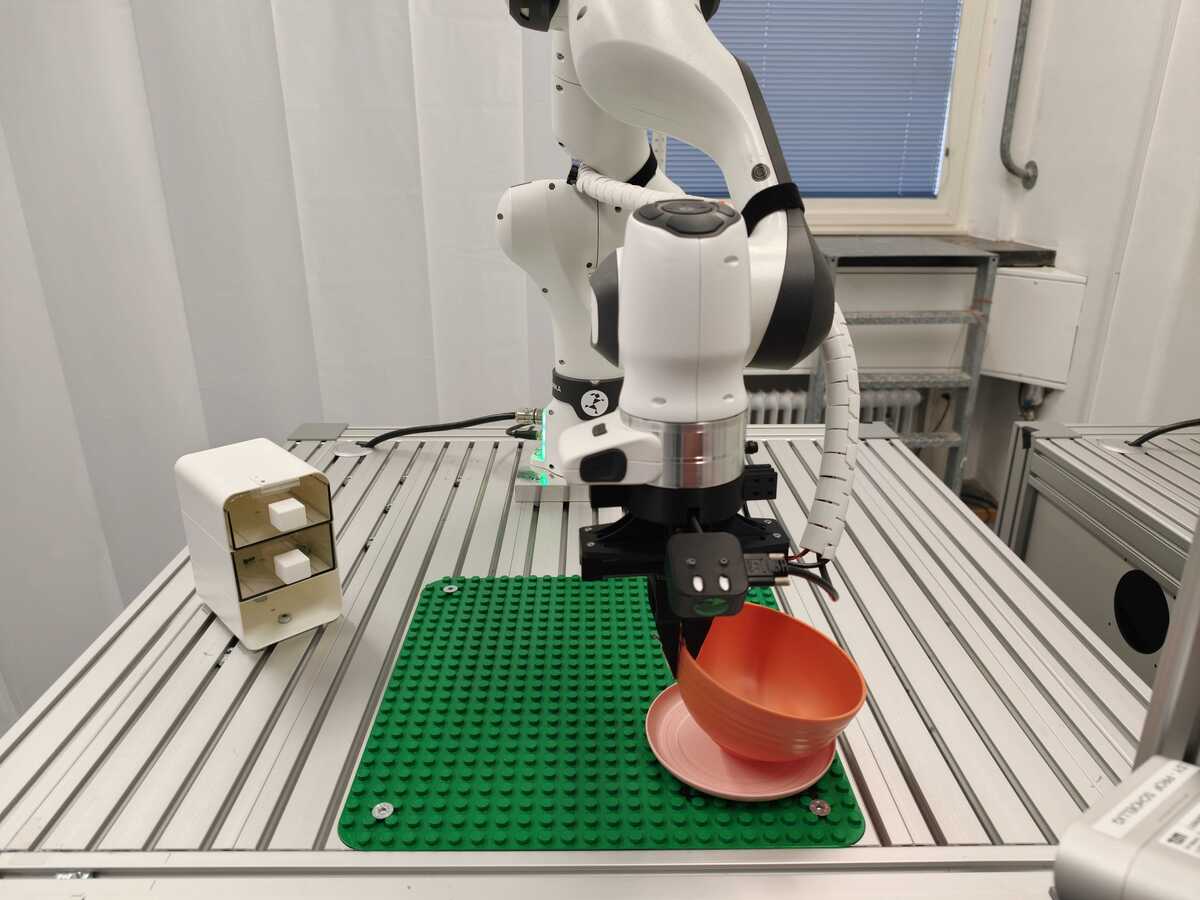}%
        \label{fig:0_put_bowl}%
    }\hfill
    \subfigure[2. \textsc{Stack}
    ]{%
        \centering
        \includegraphics[width=0.195\linewidth, trim={6cm 0cm 10cm 0cm}, clip]{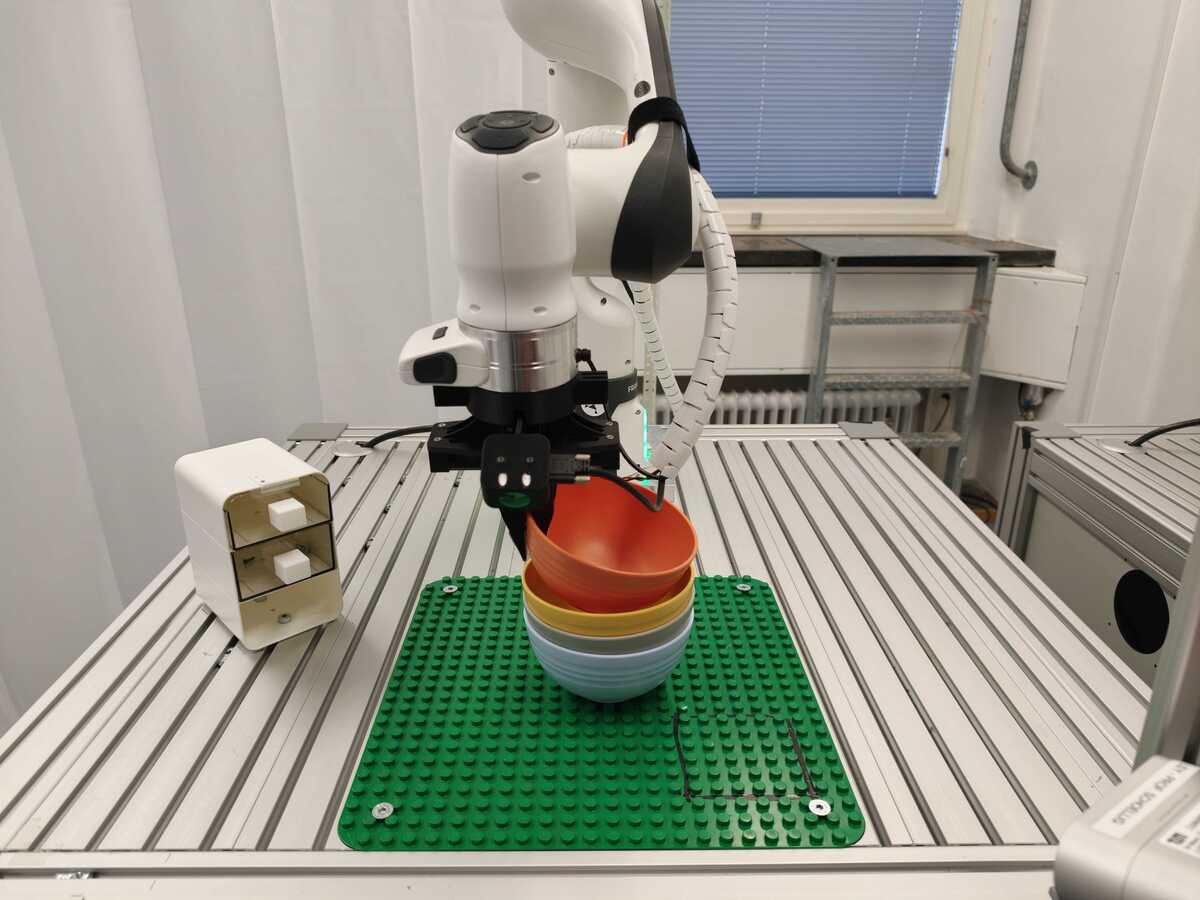}%
        \label{fig:1_stack_bowls}%
    }\hfill
    \subfigure[3. \textsc{Moka}
    ]{%
        \centering
        \includegraphics[width=0.195\linewidth, trim={6cm 0cm 10cm 0cm}, clip]{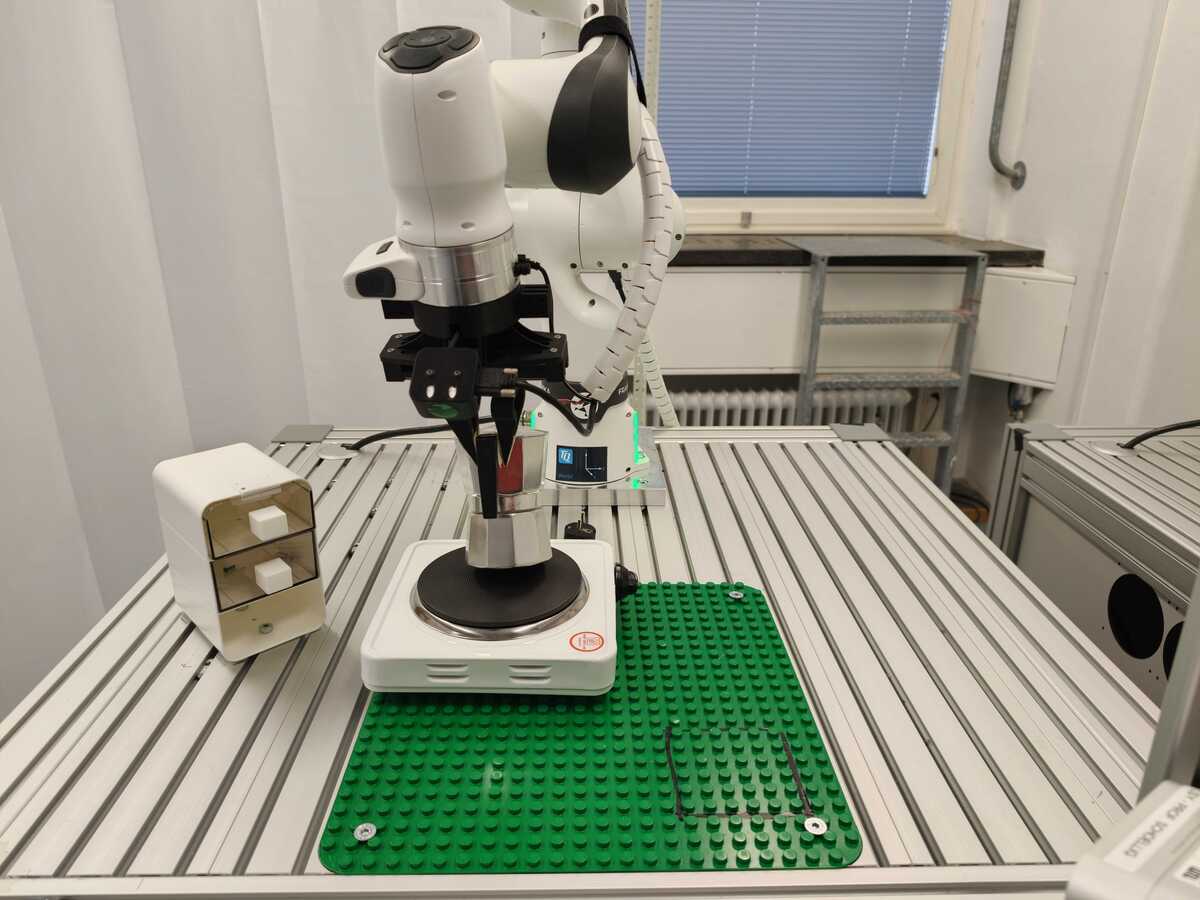}%
        \label{fig:2_put_moka}%
    }\hfill
    \subfigure[4. \textsc{Drawer}
    ]{%
        \centering
        \includegraphics[width=0.195\linewidth, trim={6cm 0cm 10cm 0cm}, clip]{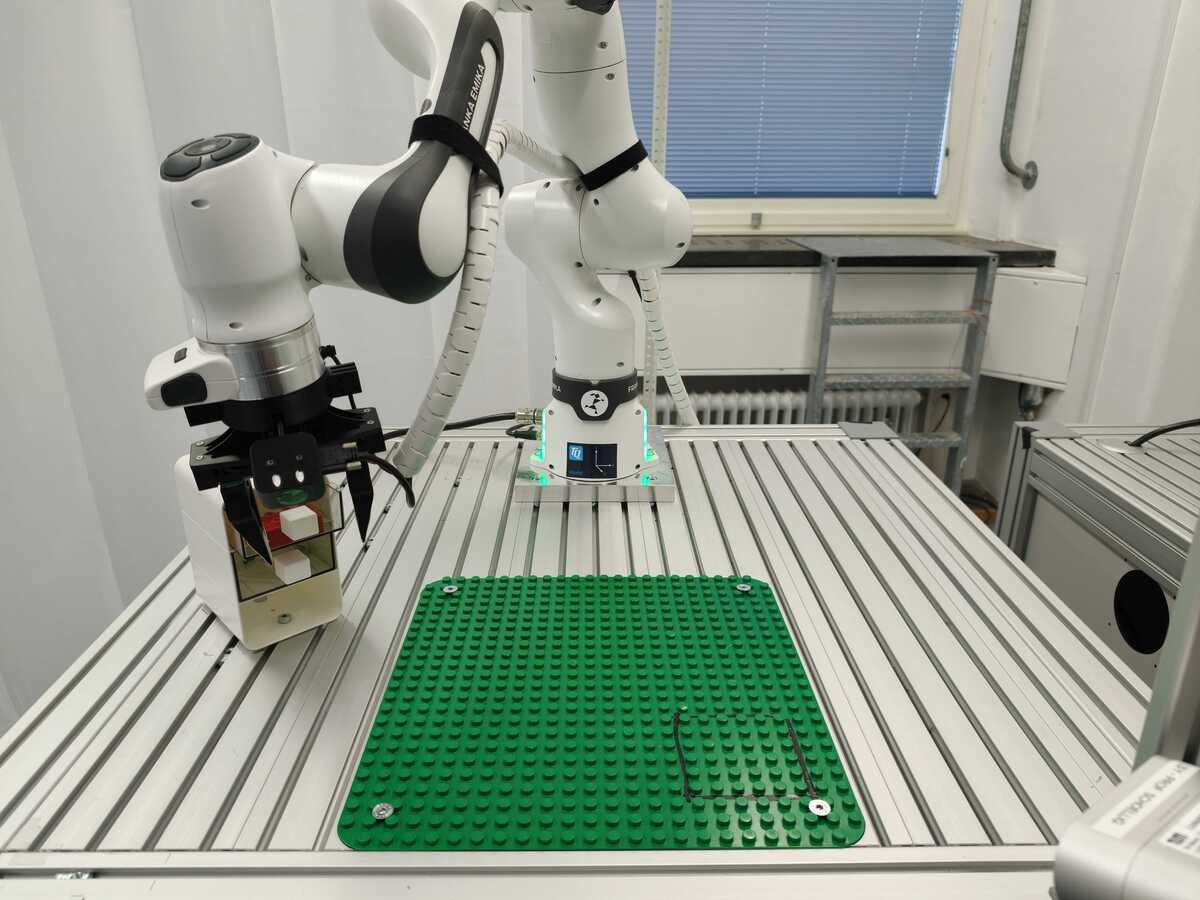}%
        \label{fig:3_close_drawer}%
    }\hfill
    \subfigure[5. \textsc{Lego}
    ]{%
        \centering
        \includegraphics[width=0.195\linewidth, trim={6cm 0cm 10cm 0cm}, clip]{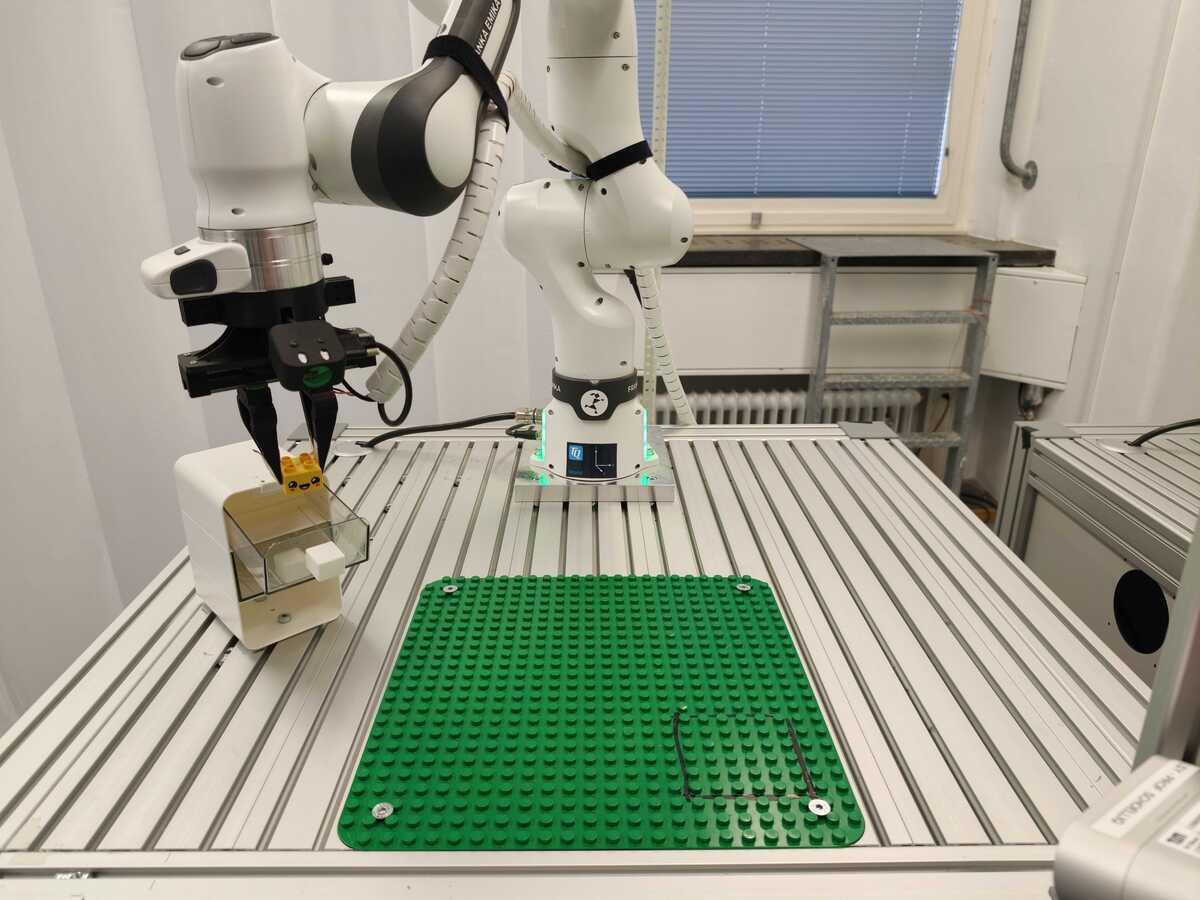}%
        \label{fig:4_put_lego}%
    }
    \hfill\\
    \vspace{-3pt}
    \caption{Our five real-world manipulation tasks involve objects of different shapes, weights, and dynamics, as well as different motion patterns.}
    \label{fig:tasks}
\end{figure}

{
\vspace{1.7mm}
\begin{table}[tb!]
\centering
\begin{tabular}{lcccc}
\toprule 
Method & AUC $\uparrow$ & FWT $\uparrow$ & NBT $\downarrow$ \\ 
\midrule
SeqFFT & 23.8 & \textbf{68.0} & 80.0\\
SeqLoRA & 22.9 & 64.0 & 76.9 \\
ER & 51.1 & 60.0 & 17.1 \\
\textbf{\fw (ours)} & \textbf{63.3} & 62.0 & \textbf{-2.9} \\
\bottomrule
\end{tabular}
\caption{Overall results in our hardware experiments.}
\label{tab:hardware_results}
\end{table}
}

\begin{table}[tb!]
\centering
\setlength{\tabcolsep}{3pt}
\newcolumntype{C}[1]{>{\centering\arraybackslash}p{#1}}
\newlength{\taskw}
\setlength{\taskw}{1.6em}
\newcommand{\taskhead}[1]{\makebox[4.6em]{\textsc{#1}}}
\begin{tabular}{l | C{\taskw}C{\taskw} | C{\taskw}C{\taskw} | C{\taskw}C{\taskw} | C{\taskw}C{\taskw} | C{\taskw}C{\taskw}}
\toprule
 & \multicolumn{2}{c|}{\taskhead{1. Bowl}} & \multicolumn{2}{c|}{\taskhead{2. Stack}} & \multicolumn{2}{c|}{\taskhead{3. Moka}} & \multicolumn{2}{c|}{\taskhead{4. Drawer}} & \multicolumn{2}{c}{\taskhead{5. Lego}} \\
Stage & \ding{192} & \ding{193}
& \ding{192} & \ding{193}
& \ding{192} & \ding{193}
& \ding{192} & \ding{193}
& \ding{192} & \ding{193} \\
\midrule
1 & \textbf{100} & \textbf{100} & -- & -- & -- & -- & -- & -- & -- & -- \\
2 & \textbf{100} & 80 & \textbf{70} & 50 & -- & -- & -- & -- & -- & -- \\
3 & \textbf{80} & \textbf{80} & 50 & \textbf{60} & \textbf{20} & \textbf{20} & -- & -- & -- & -- \\
4 & 80 & \textbf{90} & \textbf{70} & 50 & 20 & \textbf{50} & 80 & \textbf{90} & -- & -- \\
5 & 60 & \textbf{90} & 50 & \textbf{60} & 10 & \textbf{30} & 50 & \textbf{90} & 30 & \textbf{50} \\
\bottomrule
\end{tabular}
\caption{Evolution of per-task success rates [\%] across all stages in our real-world experiments. \ding{192}\,ER, \ding{193}\,CLARE.}
\label{tab:hardware_success_rates}
\end{table}

\begin{figure}[tb!]
    \centering
    \includegraphics[width=\linewidth]{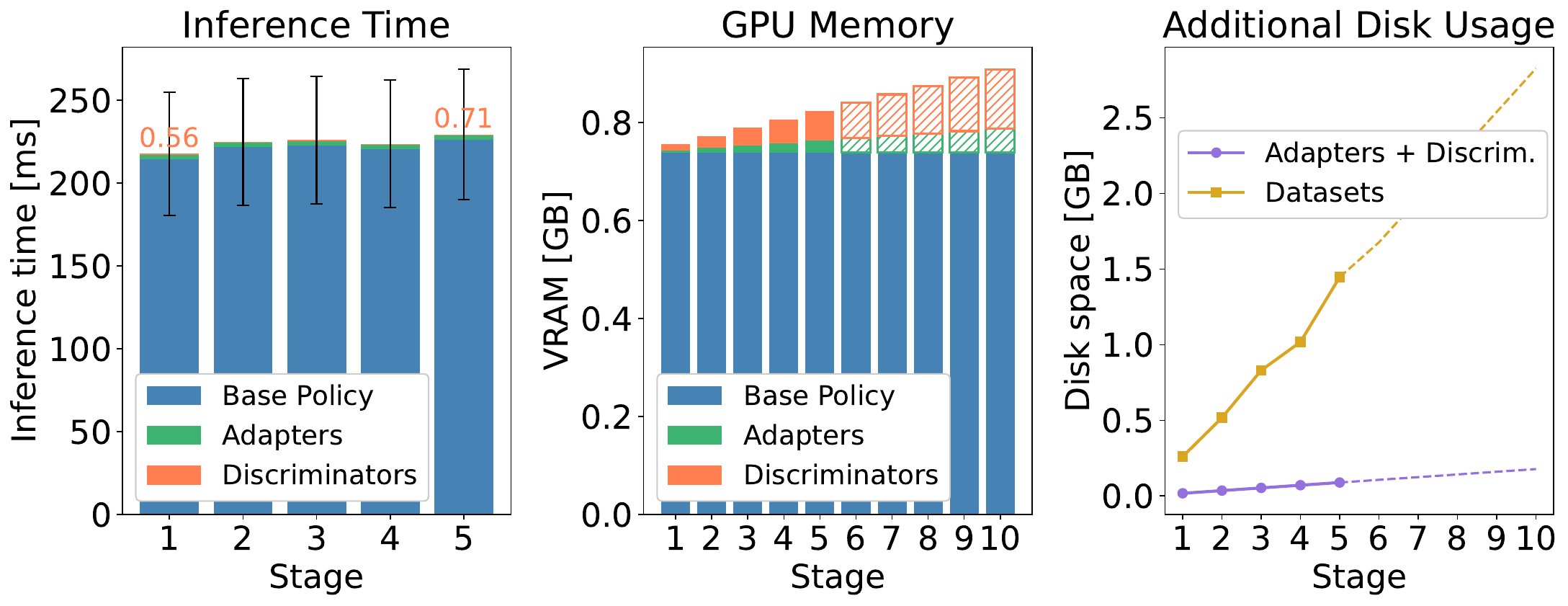}
    \caption{Inference time and memory complexity of CLARE in our hardware experiments. 
    The values for stages 6-10 are linearly extrapolated.}
    \label{fig:computation_analysis}
\end{figure}

{
\subsection{Real-World Results}
Given the importance of expanding the observation encoding modules (see Table~\ref{tab:layer_ablation}), we add adapters to the linear projection layers and the scaling and shifting AdaLN modules of each transformer decoder layer for our hardware experiments.
To keep inference latency and memory overhead low, we implement the adapters using LoRA.
Due to the scale of the hardware evaluation, we focus on three baselines: SeqFFT, SeqLoRA, and ER.
The results are provided in Tables~\ref{tab:hardware_results} and~\ref{tab:hardware_success_rates}.
CLARE achieves an AUC of 63\%, which is 12 percentage points higher than ER, and shows no catastrophic forgetting with an NBT of -2.9\%.
SeqFFT and SeqLoRA achieve high performance on new tasks, but cannot sufficiently retain the relevant representations from previous tasks.
In summary, CLARE demonstrates strong capabilities for continual learning under real-world operating conditions.

Figure~\ref{fig:computation_analysis} analyzes the inference time and memory overhead of CLARE.
The inference-time overhead relative to the base policy is below~3$\,\si{ms}$, and GPU VRAM utilization increases only by about~2\% per task.
Compared to explicitly storing data from previous tasks for ER, our approach of acquiring new skills by training dedicated lightweight modules is more storage-efficient and reduces forgetting to near zero in our experiments.
}
\section{Conclusions}
\fw enables continual learning without forgetting in VLAs, requiring neither stored exemplars nor task IDs.
By combining lightweight adapters, an autonomous expansion strategy, and an autoencoder-based routing module, our approach increases model capacity only when needed while retaining prior representations. 
Across multiple LIBERO suites, \fw achieves and maintains high task success, outperforming even strong baselines that have access to previous data.
Hardware experiments on five manipulation tasks with diverse interaction dynamics confirm these findings, demonstrating the potential of CLARE for long-term deployment of robots in non-stationary, real-world environments.

\section*{Acknowledgements}
This work was supported by the German Federal Ministry of Research, Technology and Space (BMFTR) under the Robotics Institute Germany (RIG) funded by BMFTR grant 16ME0997K, the German Research Foundation (DFG) within the RTG project ConVeY funded by grant GRK 2428, and the Humboldt Professorship for Robotics and Artificial Intelligence.


\bibliographystyle{IEEEtran}
\bibliography{ref}

\section{Appendix}
\label{sec:appendix}

\subsection{Details about the Experimental Setup}
Our five real-world tasks shown in~\Cref{fig:tasks} introduce multiple levels of diversity:
\begin{itemize}
    \item \emph{Different objects}: The tasks involve bowls, a plate, a moka pot, a stove, a drawer, and Lego blocks. Besides visual differences, these objects also have very different weights. For example, the moka pot weighs $0.5\,\si{kg}$, about 70 times as much as the Lego block. 
    \item \emph{Grasp configurations}: The tasks require different gripper widths (e.g., fully closed for the bowls and about half-closed for the Lego block) and angles (vertical for the Lego block and at a 30° angle to close the drawer) for object manipulation.
    \item \emph{Dynamic differences}: The tasks involve distinct interaction dynamics between the robot and its environment. For example, in \textsc{Moka}, the moka pot hangs at an angle when lifted by its handle, and the robot must use friction between the pot and the stove to place it upright. In \textsc{Drawer} and~\textsc{Lego}, the plastic drawer compartment exhibits a strongly nonlinear friction profile, requiring the robot to exert significant force to close the drawer. 
    \item \emph{Strategy switching}: \textsc{Lego}, for example, consists of three stages: picking up the block, placing it in the drawer, and closing the drawer.
\end{itemize}

\subsection{Additional Experiments}

\begin{figure*}[tb!]
    \vspace{1.7mm}
    \centering
    \includegraphics[width=\linewidth]{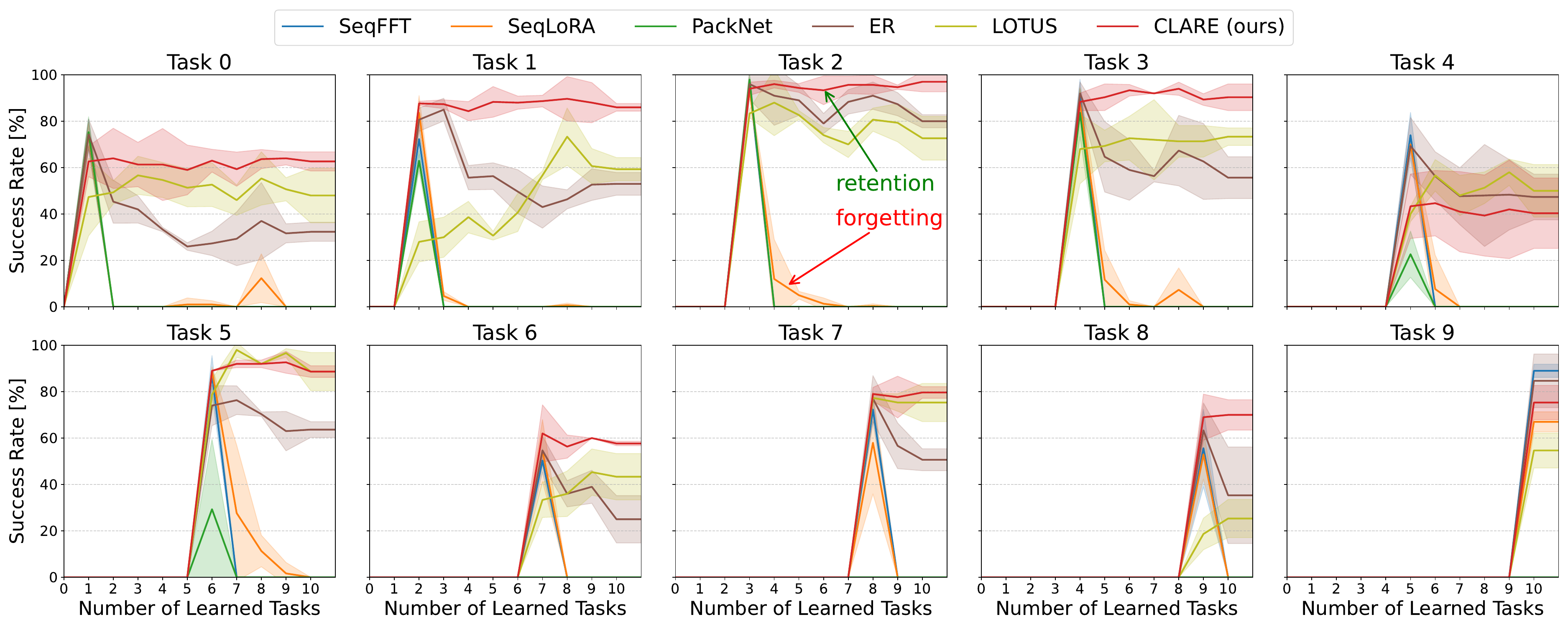}
    \caption{Success rate curves of \fw and five baselines on the LIBERO-Long benchmark. The solid lines represent the average success rates across three random seeds, and the shaded regions indicate the standard deviations. The results demonstrate that our method achieves a higher overall success rate and mitigates catastrophic forgetting more effectively during continual learning than the baselines, despite ER and LOTUS using previous data.   
    }
    \label{fig:baseline_comparison}
\end{figure*}
\Cref{fig:baseline_comparison} shows the continual learning behavior on LIBERO-Long for a selected set of baselines. These results further demonstrate that \fw does not exhibit catastrophic forgetting while achieving a high overall success rate across tasks.

\begin{figure*}[tb!]
    \centering
    \includegraphics[width=\linewidth, trim={0cm 0cm 1cm 0cm}, clip]{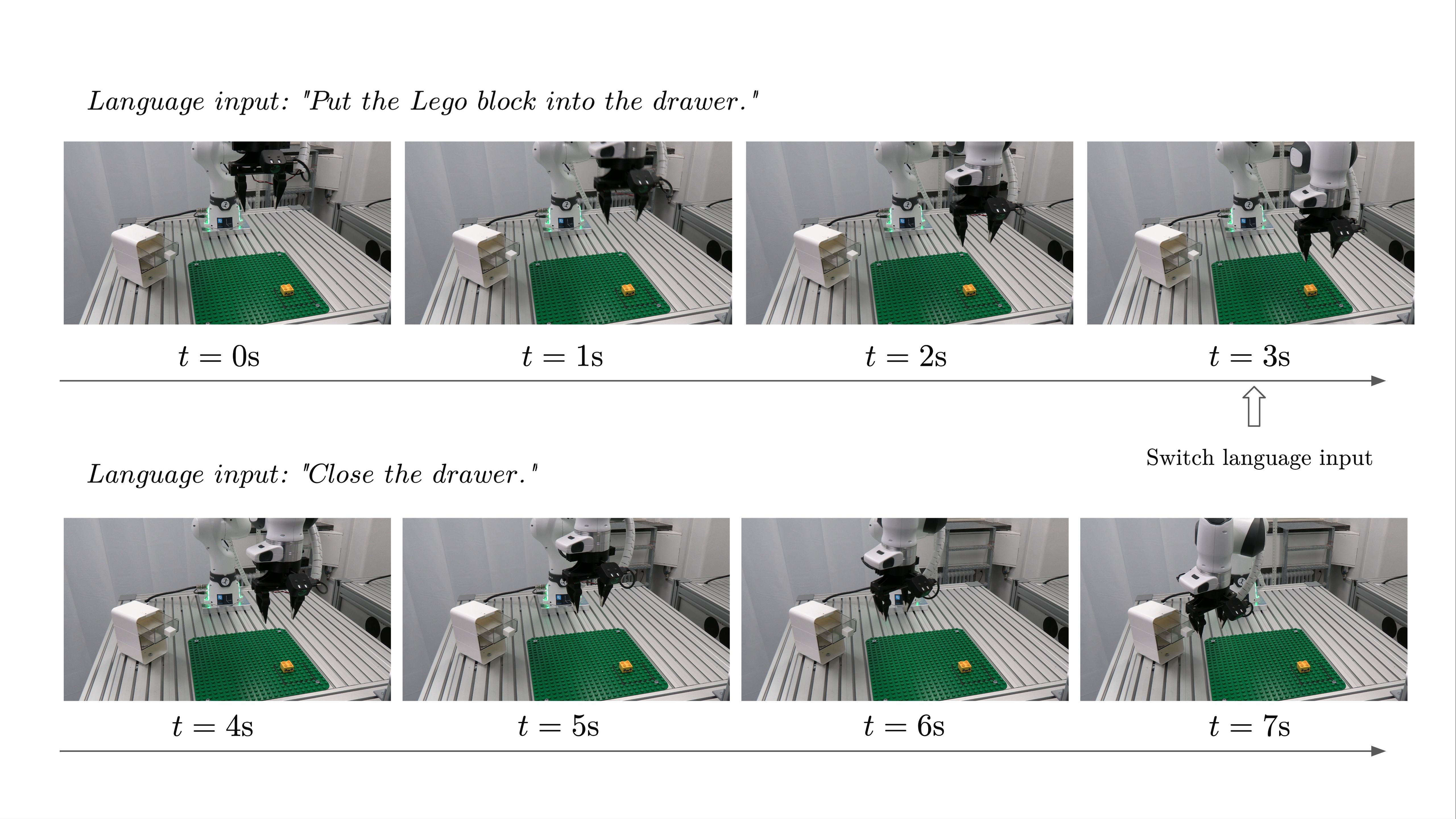}\\
    \includegraphics[width=0.6\linewidth]{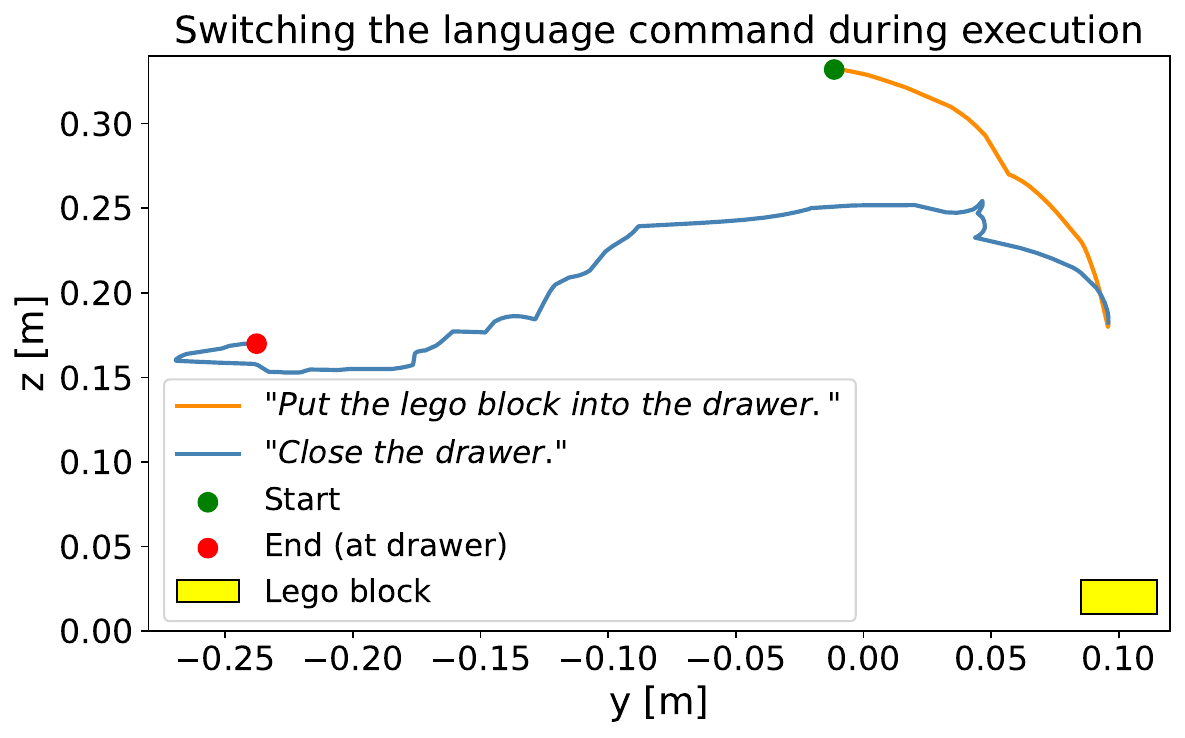} \\
    \caption{When changing the language command during execution, CLARE correctly adapts its behavior by switching the activated adapters.}
    \label{fig:switching}
\end{figure*}

In another experiment, we investigate how quickly and reliably the router responds to \textit{changes} in the active task during deployment.
To this end, we change the language command from \textit{``put the Lego block into the drawer''} to \textit{``close the drawer''} mid-execution after three seconds. 
The router immediately switches to the correct adapter; the robot stops approaching the block and moves directly toward the drawer.  
An exemplary rollout is shown in Figure~\ref{fig:switching}.
Crucially, tasks \textsc{Drawer} and \textsc{Lego} involve almost identical scenes (same workbench, same drawer), so the router must rely on subtle differences in the language and proprioceptive features to distinguish them.
This behavior was consistent across 10 rollouts, demonstrating that our reconstruction-based routing can robustly switch capabilities to adapt to dynamic context changes during deployment.
We further test the router in a continuous multi-task scenario: by changing the language prompt and rearranging the scene between tasks, CLARE successfully completes a sequence of three different tasks (\textsc{Lego} $\rightarrow$ \textsc{Bowl} $\rightarrow$ \textsc{Drawer}) without any task identifiers, activating the correct adapter for each context.

\begin{table}[tb!]
\centering
\setlength{\tabcolsep}{3pt}
\newcolumntype{C}[1]{>{\centering\arraybackslash}p{#1}}
\setlength{\taskw}{1.6em}
\newcommand{\taskhead}[1]{\makebox[4.6em]{\textsc{#1}}}
\begin{tabular}{l | C{\taskw}C{\taskw} | C{\taskw}C{\taskw} | C{\taskw}C{\taskw} | C{\taskw}C{\taskw} | C{\taskw}C{\taskw}}
\toprule
 & \multicolumn{2}{c|}{\taskhead{1. Bowl}} & \multicolumn{2}{c|}{\taskhead{2. Stack}} & \multicolumn{2}{c|}{\taskhead{3. Moka}} & \multicolumn{2}{c|}{\taskhead{4. Drawer}} & \multicolumn{2}{c}{\taskhead{5. Lego}} \\
Stage & \ding{194} & \ding{197}
& \ding{194} & \ding{197}
& \ding{194} & \ding{197}
& \ding{194} & \ding{197}
& \ding{194} & \ding{197} \\
\midrule
1 & 90 & \textbf{100} & -- & -- & -- & -- & -- & -- & -- & -- \\
2 & \textbf{100} & 80 & 20 & \textbf{50} & -- & -- & -- & -- & -- & -- \\
3 & \textbf{100} & 80 & 20 & \textbf{60} & \textbf{20} & \textbf{20} & -- & -- & -- & -- \\
4 & \textbf{90} & \textbf{90} & 40 & \textbf{50} & 40 & \textbf{50} & 60 & \textbf{90} & -- & -- \\
5 & \textbf{100} & 90 & \textbf{60} & \textbf{60} & 20 & \textbf{30} & 60 & \textbf{90} & 30 & \textbf{50} \\
\bottomrule
\end{tabular}
\caption{Evolution of per-task success rates [\%] across all stages in our real-world experiments for different numbers of expandable AdaLN modules. \ding{194}\,three modules, \ding{197}\,six modules.}
\label{tab:hardware_layer_ablation}
\end{table}

We also evaluate the impact of the number of expandable AdaLN modules in our hardware experiments and provide the results in~\Cref{tab:hardware_layer_ablation}. 
Injecting adapters into all scale and shift modules performs best.

\subsection{Extended Discussion}
In our ablation experiment (see~\Cref{tab:layer_ablation}), expanding only the encoder performs similarly to expanding both the encoder and the decoder, but much better than expanding only the decoder. Our main conclusion from these results is that the observation-conditioning modules are crucial for injecting adapters during continual learning.

Based on these findings, we expand only the modules in the conditioning pathway of the DiT-Dec architecture for our real-world experiments: 1) the linear layers projecting the observation features into the same token dimension, and 2) the scale and shift AdaLN modules, which inject the observation condition into the transformer decoder layers. Our hardware experiments involve changes in daylight, reflections, and slight drifts of the camera extrinsics. 

Despite the physical diversity of the real-world tasks, \fw achieves an AUC of$~63.3\%$ and a near-zero NBT of~$-2.9\%$ in our hardware experiments (see Table~\ref{tab:hardware_results}), significantly outperforming all baselines.
The detailed per-stage success rates in Table~\ref{tab:hardware_success_rates} further confirm that our expansion strategy preserves performance on earlier tasks even as new, physically distinct tasks are learned.
These results empirically validate that focusing the expansion on the observation-conditioning pathway is not a bottleneck for physical dynamics in our setup, supporting the conclusion drawn from the simulation ablation.
However, we note that this expansion strategy assumes that the VLA has been pretrained on a sufficiently large dataset of robot demonstrations that cover diverse motion patterns.
With the availability of large-scale robotic datasets, such as DROID~\cite{khazatsky2024droid}, we believe this prerequisite can be met for common robot embodiments.

\end{document}